# MetaboLLM: a metabolomics-specialized large language model for biochemical knowledge integration and predictive metabolite graph construction

**[1]Dohyun Ku, [2]Min Gu Kwak, [3]Francisco J. Pasquel, [1]Jing Li**

[1]H. Milton Stewart School of Industrial and Systems Engineering, Georgia Institute of Technology, Atlanta, GA

[2]Capital One, NY

[3]Department of Medicine, Division of Endocrinology, Emory University School of Medicine, Atlanta, GA

**Abstract**: Metabolomics knowledge is distributed across heterogeneous resources and remains difficult to translate into predictive representations. We developed MetaboLLM, a metabolomics-specialized large language model adapted through continual pretraining, supervised fine-tuning, and structured retrieval, together with MetaboLLM-GIN, which converts generated biochemical descriptions into metabolite graphs for patient-level prediction using a graph isomorphism network. Across four backbone families, MetaboLLM outperformed corresponding base and medically adapted models on metabolomics knowledge, relational, and description tasks, and transferred to an external public benchmark. MetaboLLM-GIN achieved the highest AUC for stress hyperglycemia prediction after coronary artery bypass grafting (0.8616) and postmenopausal hormone-regimen classification (0.8123), outperforming conventional models, alternative graph constructions, and graphs generated from unadapted or non-retrieval LLM configurations. Model interpretation further produced biologically meaningful findings in both applications. These results show that domain-specialized language models can organize heterogeneous biochemical knowledge into predictive and interpretable metabolite graph representations.

## Introduction

Metabolites provide a functional readout of cellular biochemistry, allowing metabolomics to capture molecular alterations associated with physiological and disease states[1,2]. Advances in mass spectrometry now enable the simultaneous detection and quantification of thousands of metabolite features, substantially expanding the scale of biomedical metabolomics studies[3,4]. Consequently, metabolomics has become an important approach for disease characterization and biomarker discovery, with growing applications in precision medicine, treatment monitoring, and clinical outcome prediction[5–8]. However, moving beyond isolated biomarkers toward predictive analysis requires considering the interconnected biochemical processes that give individual metabolites their biological context[9,10].

Biochemical knowledge is distributed across public resources with different scopes and data structures. KEGG provides structured pathways, reactions, and enzymes, whereas the Human Metabolome Database (HMDB) emphasizes human metabolites, biological roles, disease associations, and compound annotations[11,12]. PubChem offers broad chemical coverage, structural identifiers, molecular properties, and compound descriptions, while the Small Molecule Pathway Database (SMPDB) provides curated pathway information and physiological descriptions relevant to human metabolism[13,14]. As identifiers, naming conventions, ontologies, and annotation formats differ across them, the same metabolite is often split across several partially overlapping records. Previous efforts have addressed parts of this heterogeneity through metabolite identifier consolidation, metabolite and reaction standardization, and the integration of functional and chemical annotations for database querying and enrichment[15–18]. Those efforts were built for identifier mapping, network reconciliation, model interoperability, and functional enrichment. Developing a metabolomics-specialized language model requires information to be reorganized into a consistent representation that connects metabolite entities, chemical attributes, pathway-reaction-enzyme relationships, and textual descriptions.

Large language models (LLMs) are now applied across biomedical and clinical tasks, including medical question answering, information extraction, documentation, and diagnostic support[19,20]. Domain adaptation drives much of this progress. Med-PaLM demonstrated the value of medical instruction adaptation and evaluation of clinical knowledge[21]. PMC-LLaMA combined continual pretraining (CPT) on biomedical literature and textbooks with medical supervised fine-tuning (SFT)[22]. Meditron scaled CPT using biomedical publications and clinical guidelines[23]. Me-LLaMA further applied CPT on biomedical literature and clinical notes followed by SFT on diverse biomedical and clinical text-analysis tasks[24]. Retrieval-augmented frameworks, such as MedRAG[25] and BiomedRAG[26], improve access to external evidence across medical question answering and biomedical natural language processing tasks. All of these systems are organized around clinical text, literature, and notes as the training signal and documents or passages as the retrieval unit. None of these corresponds to a metabolite, a reaction, or an enzyme, and none was built to represent the relationships among them.

Recent studies have begun to explore language models and LLM-based systems for more specific metabolomics objectives. DeepMet learned a chemical language from the molecular structures of known metabolites to anticipate previously unrecognized metabolites[27]. MetaboliteChat integrates molecular graphs and images with a language model to generate free-form predictions of metabolite properties and mechanisms[28]. MetaboT enables natural language interaction with an existing metabolomics knowledge graph by translating user questions into queries[29]. MetaBench evaluates off-the-shelf models across knowledge, understanding, grounding, reasoning, and research[30]. Each treats the language model as a structure encoder, a query interface, or a subject of evaluation. However, it remains unclear whether a language model can integrate biochemical knowledge across databases and capture the relationships among metabolites, pathways, reactions, and enzymes. It is also unknown whether such a model can generate descriptions grounded in retrieved biochemical records and transfer to an external benchmark.

Improved performance on biochemical question answering or description generation does not establish that the LLM knowledge is useful for patient-level metabolomics analysis. Network-based methods have long organized and interpreted metabolomics knowledge for this purpose. MetaMapp combines curated reactant relationships with chemical and mass-spectral similarities[31], while FELLA propagates metabolite-level observations through a hierarchy of reactions, enzymes, modules, and pathways[32]. More recent knowledge-graph efforts, including MetaKG and MetaboKG, support the integration, querying, and reuse of metabolomics knowledge and analytical evidence, but they are designed as knowledge infrastructure rather than patient-level outcome prediction[33,34]. Metabolitics maps biofluid measurements onto a genome-scale metabolic network to estimate patient-specific reaction and pathway activities[35], whereas M-GNN represents patients, metabolites, pathways, and diseases in a heterogeneous graph for lung cancer classification[36]. MP-GNN connects metabolites by documented biochemical interactions and pathway co-membership for patient-level risk prediction, but its topology remains defined by curated annotations[37]. PathwaySeeker connected metabolic graphs with evidence-grounded LLM reasoning, although the graph serves as context for mechanistic interpretation rather than being derived from the model and evaluated for classification[38].

Graph-based learning connects biochemical knowledge to quantitative profiles by representing measured metabolites as nodes, their relationships as edges, and sample-specific abundances as node features. Curated pathway and reaction graphs provide explicit biological relationships but may offer limited direct connectivity among the metabolites measured and annotated in a particular dataset, whereas correlation and partial-correlation networks can recover broader data-driven associations but produce cohort-dependent topologies influenced by sample size, regularization, and network-construction choices[39–41]. It remains unclear whether cross-database biochemical knowledge acquired by a metabolomics-specialized LLM can be converted into a metabolite topology that improves patient-level prediction beyond curated pathway, reaction, and cohort-derived correlation graphs.

To address these challenges, we developed MetaboLLM, a metabolomics-specialized language model, and MetaboLLM-GIN, a framework that translates its biochemical knowledge into graph-based predictive representations. Records from KEGG, HMDB, PubChem, and SMPDB were harmonized into a unified resource connecting metabolite identities and chemical attributes with pathways, reactions, enzymes, and textual descriptions. MetaboLLM was adapted through metabolomics-specific CPT and SFT, supported by a newly constructed instruction and evaluation benchmark. Structured retrieval further provided reaction- and pathway-level context to complement knowledge acquired during model adaptation. To extend this knowledge beyond language tasks, semantic similarity among MetaboLLM-generated biochemical descriptions was used to construct a knowledge-informed metabolite topology. Sample-specific metabolite measurements were then assigned as node features, enabling a graph isomorphism network (GIN) to combine biochemical context with quantitative profiles for patient-level classification.

The framework was evaluated across three complementary levels. First, metabolomics-specific factual and relational capabilities were compared across general-purpose, medically adapted, and MetaboLLM models, while adaptation across multiple backbone families was examined for consistency. Second, metabolite, pathway, and enzyme description generation and the contribution of retrieved biochemical context were evaluated using the metabolomics benchmark, while external transfer was assessed on MetaBench, an externally developed public metabolomics benchmark. Third, the downstream utility of the language-derived graph was examined in metabolomics applications. MetaboLLM-GIN was compared with several competing models, as well as pathway-, reaction-, correlation-, and random-graph alternatives under a common GIN architecture. LLM-derived graphs generated by the unadapted backbone and MetaboLLM, each with and without retrieval, were also compared. Influential edges were compared with curated biochemical relationships, and important metabolites were examined through pathway-level analysis to investigate the biological coherence of the predictive representations. These evaluations connect heterogeneous knowledge integration and domain-specific language modeling with knowledge-informed graph construction, patient-level metabolomics analysis, and biological interpretation.

## Results

### Overview of MetaboLLM framework

Fig. 1 summarizes the MetaboLLM framework, which connects cross-database biochemical knowledge integration, natural language corpus construction, domain-specific model adaptation, language model evaluation, and downstream graph-based prediction. Records from KEGG, HMDB, PubChem, and SMPDB were harmonized into a unified metabolomics resource and converted into training samples for CPT and SFT. The resulting MetaboLLM models were evaluated on internal metabolomics tasks and an external public benchmark. Retrieval-grounded metabolite descriptions were subsequently used to construct a knowledge-informed graph for patient-level prediction using MetaboLLM-GIN.

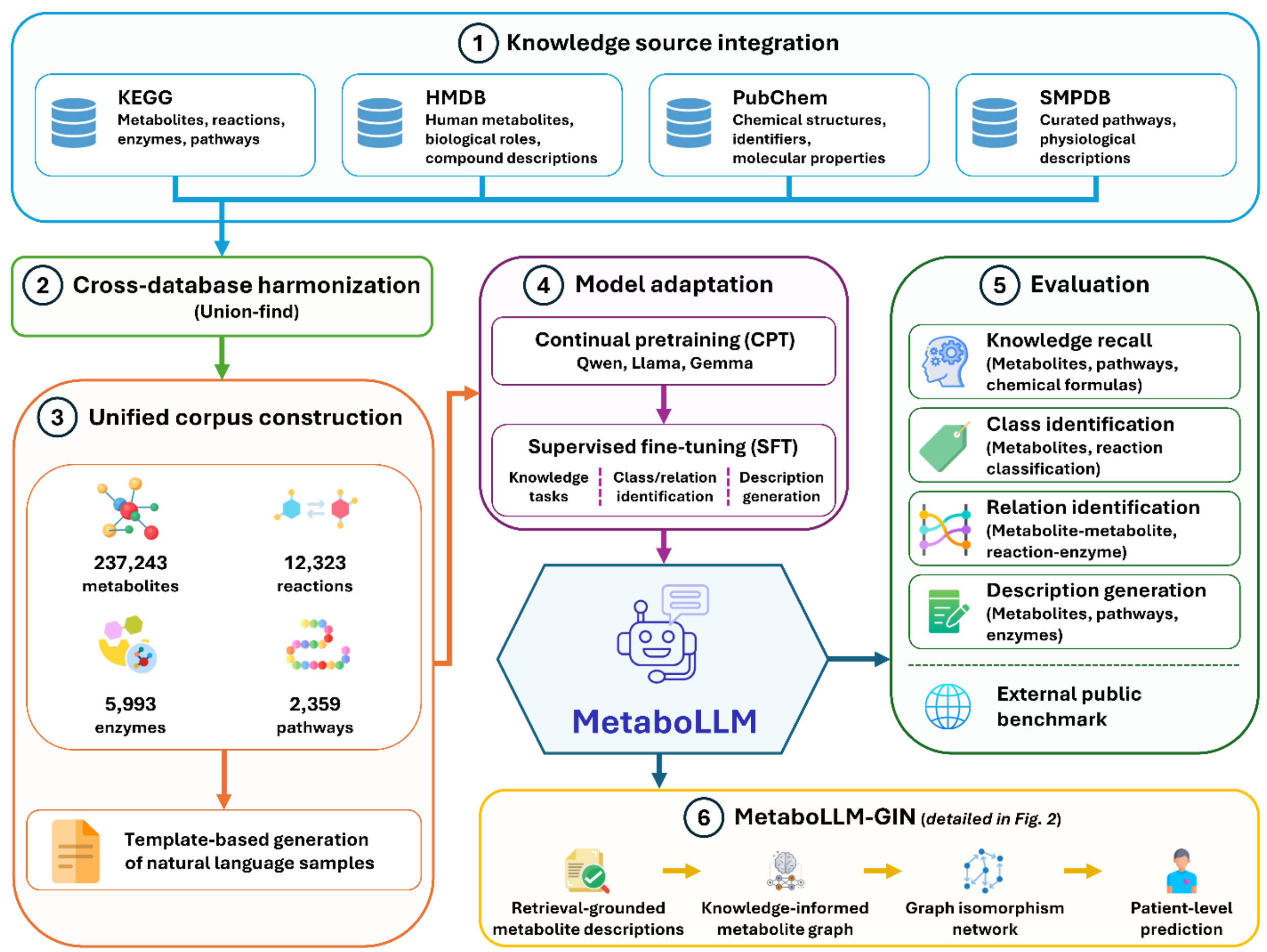

**Fig. 1 | Development and evaluation of the MetaboLLM framework.** (1) Metabolomics knowledge was collected from KEGG, HMDB, PubChem, and SMPDB. (2) Records from these databases were harmonized across shared entities and relationships. (3) The harmonized resource, comprising 237,243 metabolites, 12,323 reactions, 5,993 enzymes, and 2,359 pathways, was converted into template-based natural language samples to construct a unified corpus. (4) MetaboLLM was developed through continual pretraining (CPT) followed by supervised fine-tuning (SFT) on knowledge recall, class identification, relation identification, and description generation tasks. (5) The adapted model was evaluated across these four internal task categories and an external public benchmark. (6) Retrieval-grounded metabolite descriptions generated by MetaboLLM were transformed into a knowledge-informed metabolite graph and integrated with sample-specific metabolite profiles using a graph isomorphism network (GIN) for patient-level prediction.

### Construction of a unified metabolomics knowledge corpus

We constructed a unified metabolomics knowledge corpus by integrating four public databases: KEGG, HMDB, PubChem, and SMPDB. The resulting corpus contains 237,243 metabolites, 2,359 pathways, 12,323 reactions, and 5,993 enzymes. Each source contributed distinct knowledge types. KEGG provided structured biochemical relationships, including compound-level annotations, reaction definitions, and pathway maps. HMDB contributed human metabolite annotations, including biological roles, chemical taxonomy, disease relevance, and detailed compound descriptions. PubChem expanded chemical coverage through molecular structure information, computed properties, and compound descriptions, particularly for compounds incompletely annotated in metabolism-focused databases. SMPDB complemented pathway-level information by providing narrative descriptions of small-molecule pathways and their physiological relevance. Cross-database alignment linked metabolite, pathway, reaction, and enzyme records into unified entries while preserving source-specific annotations. These records were then converted into natural language samples covering compound, pathway, and reaction knowledge for CPT.

**Construction of metabolomics instruction and evaluation tasks**

We built a metabolomics instruction and evaluation set organized into four task categories: knowledge recall, class identification, relation identification, and description generation. This design allowed the evaluation to cover both closed-form factual prediction and long-form biochemical explanation. Closed-form tasks were formulated as short-answer or multiple-choice questions. Knowledge recall tasks evaluated compound identity and pathway membership, including name-to-formula prediction, formula-to-name prediction, metabolite-to-pathway prediction, and pathway-to-metabolite prediction. Class identification tasks evaluated recognition of metabolite and reaction class labels. Relation identification tasks evaluated structured biochemical associations among metabolites, reactions, and enzymes, including shared-reaction metabolite prediction, reaction-to-enzyme prediction, and reaction–enzyme pair validation.

Description tasks covered metabolites, pathways, and enzymes. Two additional metabolite-description regimes were constructed to represent separated annotation settings. Structure-rich metabolites were paired with retrieved reaction and pathway context, whereas structure-poor metabolites were described without external retrieval context and therefore relied on parametric biochemical knowledge. These regimes were designed to reflect sparsely annotated long-tail metabolites.

Each label denotes the input and output entity types of the corresponding task. For example, NF denotes name-to-formula prediction, MPath denotes metabolite-to-pathway prediction, and REnz denotes reaction-to-enzyme prediction. Table 1 summarizes all labels, formats, and sample counts, and Supplementary Information B gives task definitions, prompt templates, distractor construction, and split procedures.

**Table 1.** Overview of MetaboLLM instruction and evaluation tasks.

| Category | Task label | Full task name | Format | Train / Test |
|---|---|---|---|---|
| Knowledge recall | NF | Name → Formula | MCQA | 200 / 200 |
| | NF-SA | Name → Formula | Short answer | 200 / 200 |
| | FN | Formula → Name | MCQA | 200 / 200 |
| | MPath | Metabolite → Pathway | MCQA | 200 / 200 |
| | PathM | Pathway → Metabolite | MCQA | 200 / 200 |
| Class identification | MClass | Metabolite → Class | MCQA | 400 / 400 |
| | ClassM | Class → Metabolite | MCQA | 100 / 100 |
| | RClass | Reaction → Class | MCQA | 400 / 400 |
| | ClassR | Class → Reaction | MCQA | 100 / 100 |
| Relation identification | MRxn | Metabolite → Reaction-linked metabolite | MCQA | 400 / 400 |
| | REnz | Reaction → Enzyme | MCQA | 300 / 300 |
| | REPair | Reaction–Enzyme Pair Validation | MCQA | 300 / 300 |
| Description generation | MDesc | Metabolite Description | Long answer | 500 / 500 |
| | PathDesc | Pathway Description | Long answer | 300 / 300 |
| | EnzDesc | Enzyme Description | Long answer | 200 / 200 |
| | RichDesc | Structure-rich Metabolite Description | Long answer | 1000 / 100 |
| | PoorDesc | Structure-poor Metabolite Description | Long answer | 1000 / 100 |

**MetaboLLM evaluation protocol**

We evaluated 17 models across four groups. The first group included five publicly available medical models adapted to general clinical or biomedical text: Qwen2.5-Aloe-Beta-7B[42], II-Medical-8B[43],

Meditron3-Qwen2.5-7B[44], MedGemma-1.5-4B[45], and FineMedLM-O1[46]. These models assessed whether general medical-domain adaptation transfers to metabolomics-specific biochemical tasks. The second group comprised four unadapted backbone models: Qwen3-4B[47], Qwen3-8B[47], Gemma-3-4B[48], and Llama-3.2-3B[49]. The third group consisted of the same four backbones after CPT but before SFT, allowing the contribution of CPT to be evaluated separately. The fourth group consisted of the corresponding MetaboLLM models obtained through CPT followed by SFT. MetaboLLM-Qwen3-4B served as the primary model, while the same two-stage adaptation was applied to the other three backbones to assess consistency across model families.

All models used greedy decoding with a temperature of zero. To reduce failures caused by noncompliance with the requested answer format, the assistant response was prefixed with "The answer is: (" for multiple-choice tasks and "The answer is:" for short-answer tasks, with generation limited to 100 new tokens. Multiple-choice and short-answer tasks were evaluated using accuracy, with molecular-formula normalization applied to the short-answer formula task. Long-form descriptions were evaluated using BERTScore-F1[50], ROUGE-L-F1[51], and BLEU-2[52], reported on a 0–100 scale.

**Biochemical knowledge and structured relationships**

MetaboLLM-Qwen3-4B achieved the highest mean accuracy across the 12 multiple-choice and short-answer tasks, reaching an average of 79.07%. Table 2 reports the per-task accuracies, with the average across tasks. Each MetaboLLM variants outperformed its corresponding CPT-only and unadapted backbone model, and all four variants outperformed every evaluated medical models. CPT-only models consistently showed intermediate performance between the corresponding unadapted backbones and MetaboLLM models. The gains ranged from +9.2 for Qwen3-8B to +23.1 for Gemma-3-4B, with the largest improvements observed for Gemma-3-4B and Llama-3.2-3B, which had the lowest initial performance.

For Qwen3-4B, the largest improvement occurred in short-answer molecular-formula generation, from 3.5 to 54.5, indicating that the standardized answer prefix alone could not elicit exact formulas. Performance also increased on tasks involving biochemical relationships, including metabolite–reaction association from 72.3 to 90.0 and reaction–enzyme pair validation from 86.3 to 97.0. Gains were smaller where the unadapted model was already near ceiling, such as in reaction-to-enzyme prediction (97.3 to 98.3).

Molecular identity mapping remained the most difficult task type, with name-to-formula and formula-to-name accuracy remaining below that of most other tasks after adaptation. All five medical models performed below both unadapted Qwen3 models and every MetaboLLM variant, suggesting that general clinical and biomedical adaptation alone did not provide comparable performance on metabolomics-specific knowledge and relationship tasks.

**Table 2.** Accuracy of evaluated models across biochemical knowledge and structured relationship tasks

| Category | Model | NF | NF-SA | FN | MPath | PathM | MClass | ClassM | RClass | ClassR | MRxn | REnz | REPair | Mean |
|---|---|---|---|---|---|---|---|---|---|---|---|---|---|---|
| Medical | MedGemma-1.5-4B | 34.5 | 1.5 | 25.0 | 68.0 | 49.5 | 71.8 | 53.0 | 70.3 | 40.0 | 40.5 | 88.3 | 45.7 | 49.0 |
| | FineMedLM-O1 | 43.5 | 2.0 | 33.0 | 87.0 | 61.5 | 79.0 | 63.0 | 76.5 | 41.0 | 59.8 | 96.0 | 53.0 | 57.9 |
| | II-Medical-8B-1706 | 44.5 | 1.5 | 29.5 | 74.0 | 67.5 | 77.5 | 74.0 | 76.8 | 62.0 | 68.3 | 93.0 | 87.7 | 63.0 |
| | Qwen2.5-Aloe-Beta-7B | 38.5 | 0.5 | 27.0 | 88.0 | 74.5 | 79.8 | 68.0 | 75.8 | 63.0 | 71.0 | 98.0 | 82.7 | 63.9 |
| | Meditron3-Qwen2.5-7B | 42.0 | 1.5 | 27.5 | 92.0 | 74.0 | 80.8 | 71.0 | 77.3 | 57.0 | 79.5 | 98.0 | 88.0 | 65.7 |
| Base | Llama-3.2-3B | 34.5 | 1.0 | 29.0 | 58.5 | 57.5 | 72.5 | 74.0 | 62.8 | 42.0 | 35.8 | 93.0 | 57.0 | 51.5 |
| | Gemma-3-4B | 35.0 | 2.0 | 24.0 | 61.5 | 57.5 | 76.0 | 59.0 | 71.3 | 41.0 | 44.0 | 90.7 | 64.3 | 52.2 |
| | Qwen3-4B | 41.5 | 3.5 | 27.0 | 91.0 | 77.0 | 79.0 | 74.0 | 80.8 | 68.0 | 72.3 | 97.3 | 86.3 | 66.5 |
| | Qwen3-8B | 42.0 | 3.0 | 29.0 | 79.5 | 80.0 | 80.5 | 78.0 | 85.3 | 64.0 | 70.0 | 96.7 | 93.3 | 66.8 |
| CPT only | Llama-3.2-3B (CPT) | 35.5 | 21.0 | 24.0 | 73.5 | 62.5 | 77.5 | 69.0 | 61.0 | 47.0 | 61.5 | 91.0 | 38.0 | 55.1 |
| | Gemma-3-4B (CPT) | 40.0 | 41.5 | 31.0 | 58.5 | 49.0 | 78.0 | 58.0 | 70.8 | 47.0 | 52.5 | 95.0 | 50.3 | 56.0 |
| | Qwen3-4B (CPT) | 51.0 | 34.5 | 30.0 | 77.0 | 74.5 | 83.0 | 79.0 | 77.0 | 60.0 | 72.5 | 97.7 | 95.3 | 69.3 |
| | Qwen3-8B (CPT) | 42.0 | 52.0 | 32.0 | 82.5 | 74.0 | 81.3 | 78.0 | 83.8 | 70.0 | 68.0 | **98.3** | 93.0 | 71.2 |
| MetaboLLM | MetaboLLM-Llama-3.2-3B | 41.0 | 37.5 | 31.5 | 94.0 | 81.0 | 83.8 | 76.0 | 80.8 | 56.0 | 84.0 | 96.0 | 78.7 | 70.0 |
| | MetaboLLM-Gemma-3-4B | 42.5 | 41.0 | 36.0 | 92.5 | **88.5** | 86.0 | **80.0** | 84.5 | 69.0 | **91.0** | 97.7 | 94.7 | 75.3 |
| | MetaboLLM-Qwen3-8B | 51.5 | 45.5 | 33.5 | 94.0 | 84.0 | 85.5 | **80.0** | 86.8 | 73.0 | 83.3 | 97.7 | 96.3 | 75.9 |
| | MetaboLLM-Qwen3-4B | **53.0** | **54.5** | **41.0** | **96.0** | 87.5 | **90.5** | 77.0 | **89.0** | **75.0** | 90.0 | **98.3** | **97.0** | **79.1** |

Mean is the average across 12 tasks; Best per column in bold; Task abbreviations are defined in Table 1.

**Biochemical description generation**

We evaluated free-text generation across metabolite, pathway, and enzyme description tasks, together with structure-rich and structure-poor metabolite subsets. Table 3 reports BERTScore-F1 for all 17 models, with ROUGE-L-F1 and BLEU-2 provided in Supplementary Tables 1 and 2.

MetaboLLM-Qwen3-4B outperformed all unadapted backbone, CPT-only, and medical models across the three standard description tasks and achieved the highest BERTScore-F1 on all five tasks. The largest improvement occurred for metabolite descriptions, where two-stage adaptation increased BERTScore-F1 from 82.39 to 91.71, ROUGE-L-F1 from 16.56 to 58.65, and BLEU-2 from 9.13 to 52.22.

Pathway and enzyme descriptions improved similarly. MetaboLLM-Qwen3-4B reached BERTScore-F1 of 87.79 on structure-rich and 88.06 on structure-poor metabolites. As the two subsets contain different compounds, these values do not isolate the effect of annotation richness, and we report them to show that adaptation did not degrade on sparsely annotated metabolites.

**Table 3.** BERTScore-F1 across biochemical description generation tasks

| Category | Model | MDesc | PathDesc | EnzDesc | RichDesc | PoorDesc |
|---|---|---|---|---|---|---|
| Medical | MedGemma-1.5-4B | 81.05 | 80.97 | 79.33 | 84.89 | 84.73 |
| | FineMedLM-O1 | 83.05 | 82.87 | 81.22 | 84.10 | 83.96 |
| | II-Medical-8B-1706 | 82.71 | 82.94 | 81.60 | 86.37 | 85.84 |
| | Qwen2.5-Aloe-Beta-7B | 83.34 | 83.39 | 81.68 | 85.47 | 84.52 |
| | Meditron3-Qwen2.5-7B | 83.69 | 83.14 | 81.99 | 84.47 | 83.76 |
| Base | Llama-3.2-3B | 82.81 | 82.56 | 81.40 | 84.56 | 84.50 |
| | Gemma-3-4B | 81.74 | 81.54 | 80.37 | 85.48 | 85.07 |
| | Qwen3-4B | 82.39 | 81.85 | 80.68 | 86.48 | 85.80 |
| | Qwen3-8B | 82.30 | 82.28 | 80.79 | 86.20 | 85.61 |
| CPT only | Llama-3.2-3B (CPT) | 84.34 | 84.30 | 81.17 | 81.62 | 82.37 |
| | Gemma-3-4B (CPT) | 84.59 | 83.13 | 81.2 | 83.22 | 83.00 |
| | Qwen3-4B (CPT) | 85.54 | 83.06 | 81.46 | 86.06 | 85.24 |
| | Qwen3-8B (CPT) | 84.86 | 82.97 | 81.21 | 85.23 | 84.80 |
| MetaboLLM | MetaboLLM-Llama-3.2-3B | 90.93 | 86.74 | 83.78 | 87.13 | 87.65 |
| | MetaboLLM-Gemma-3-4B | 87.39 | 83.83 | 80.69 | 87.03 | 87.74 |

| | MetaboLLM-Qwen3-8B | 90.94 | 87.26 | 83.70 | 87.66 | 88.02 |
|---|---|---|---|---|---|---|
| | MetaboLLM-Qwen3-4B | **91.71** | **88.04** | **84.12** | **87.79** | **88.06** |

Best per column in bold; Task abbreviations are defined in Table 1.

**Effect of retrieval augmentation**

We assessed the contribution of retrieval using a within-compound ablation in which the same structure-rich metabolites were described with and without retrieved reaction- and pathway-level context. Table 4 reports the comparison between MetaboLLM-Qwen3-4B and its corresponding unadapted backbone.

Retrieval improved all three metrics for both models. For MetaboLLM-Qwen3-4B, BERTScore-F1 rose from 86.03 to 87.79, ROUGE-L-F1 from 21.98 to 26.42, and BLEU-2 from 24.82 to 29.25. As the lexical gains were roughly twice those of the unadapted backbone, adaptation made the retrieved context more usable. MetaboLLM without retrieval still exceeded the backbone with retrieval on BLEU-2 (24.82 against 21.54), indicating that retrieval supplemented rather than replaced the knowledge acquired through two-stage adaptation.

**Table 4.** Effect of retrieval augmentation on structure-rich metabolite descriptions

| Model | Retrieval | BERT-F1 | ROUGE-L | BLEU-2 |
|---|---|---|---|---|
| Qwen3-4B (base) | X | 85.22 | 19.83 | 19.88 |
| | O | <u>86.48</u> | <u>22.11</u> | 21.54 |
| MetaboLLM-Qwen3-4B | X | 86.03 | 21.98 | <u>24.82</u> |
| | O | **87.79** | **26.42** | **29.25** |

Best per column in bold, second best underlined.

**External benchmark evaluation**

We tested all 17 models on MetaBench, an independently developed public benchmark for metabolomics[30], using its Knowledge task (multiple-choice question-answering) and Understanding task (pathway-description generation). The results are reported in Table 5.

On the Knowledge task, CPT-only models remained slightly below their corresponding unadapted backbones, whereas the addition of SFT improved accuracy across all four model families. MetaboLLM-Qwen3-8B achieved the highest accuracy at 56.42%, followed by MetaboLLM-Qwen3-4B at 52.18%. The improvements over the corresponding unadapted backbones were smaller than those observed for description generation, suggesting that the unadapted models already captured part of the factual knowledge required for multiple-choice recall.

The gains achieved by CPT-only models over their corresponding unadapted backbones across all three generation metrics were further amplified by SFT, making the MetaboLLM models the strongest overall performers. MetaboLLM-Qwen3-4B ranked first across all three generation metrics, reaching 85.19 in BERTScore-F1, 25.83 in ROUGE-L-F1, and 17.61 in BLEU-2, followed by MetaboLLM-Qwen3-8B. The original MetaBench report placed pathway-description BERTScore-F1 between roughly 81 and 84, including frontier models such as GPT-5 and Claude Sonnet 4[30]. MetaboLLM-Qwen3-4B exceeded that range, although those values were obtained under the benchmark authors' prompting and decoding settings rather than ours. As MetaBench was also built from public metabolomics resources, these results demonstrate external benchmark transfer rather than independence from the underlying knowledge sources.

**Table 5.** Performance on MetaBench, an external metabolomics benchmark

| Category | Model | Knowledge MCQA | Pathway Description | | |
|---|---|---|---|---|---|
| | | Accuracy | BERT-F1 | ROUGE-L | BLEU-2 |
| Medical | MedGemma-1.5-4B | 41.22 | 78.03 | 11.47 | 4.72 |
| | FineMedLM-O1 | 42.94 | 82.43 | 13.87 | 7.50 |
| | II-Medical-8B-1706 | 50.22 | 83.17 | 16.33 | 9.87 |
| | Qwen2.5-Aloe-Beta-7B | 48.42 | 83.52 | 15.46 | 10.04 |
| | Meditron3-Qwen2.5-7B | 50.54 | 83.38 | 17.55 | 10.47 |
| Base | Llama-3.2-3B | 44.14 | 82.59 | 15.41 | 8.50 |
| | Gemma-3-4B | 46.02 | 80.16 | 11.60 | 5.16 |
| | Qwen3-4B | 51.42 | 81.54 | 14.85 | 6.61 |
| | Qwen3-8B | 52.26 | 81.51 | 15.63 | 6.29 |
| CPT only | Llama-3.2-3B (CPT) | 43.90 | 82.93 | 16.91 | 10.32 |
| | Gemma-3-4B (CPT) | 44.30 | 82.19 | 14.45 | 7.93 |
| | Qwen3-4B (CPT) | 49.90 | 83.20 | 16.64 | 9.22 |
| | Qwen3-8B (CPT) | 52.02 | 82.10 | 15.87 | 7.15 |
| MetaboLLM | MetaboLLM-Llama-3.2-3B | 47.62 | 84.78 | 24.65 | 16.61 |
| | MetaboLLM-Gemma-3-4B | 50.70 | 82.31 | 15.47 | 7.97 |
| | MetaboLLM-Qwen3-8B | **56.42** | 85.01 | 24.59 | 16.98 |
| | MetaboLLM-Qwen3-4B | 52.18 | **85.19** | **25.83** | **17.61** |

Best per column in bold.

**Overview of MetaboLLM-GIN framework**

Fig. 2 summarizes the MetaboLLM-GIN framework, which converts language-encoded biochemical knowledge into graph representations for patient-level metabolomics prediction. For each measured metabolite, MetaboLLM generated a retrieval-grounded description integrating biological roles, reaction context, and pathway context. The resulting descriptions were embedded and compared using cosine similarity, and the top-ranked metabolite pairs were selected to define a shared knowledge-informed graph topology. Sample-specific metabolite abundances were assigned as node features, allowing a GIN to learn predictive representations for downstream classification. Important edges and metabolites were subsequently examined through curated biochemical relationships, randomized graph testing, pathway enrichment, and biological interpretation.

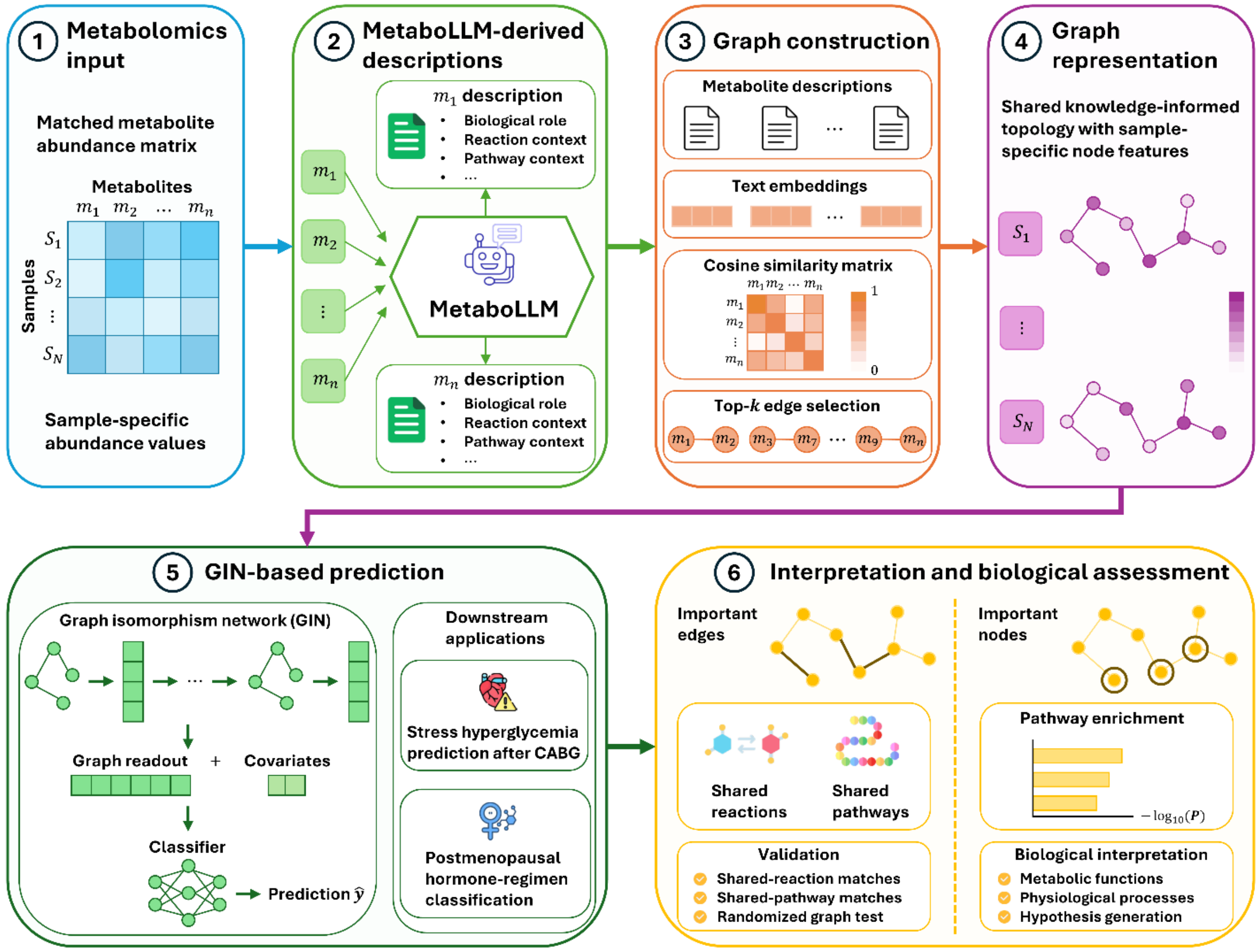

**Fig. 2 | Development and evaluation of the MetaboLLM-GIN framework.** (1) Matched metabolite abundance matrices were obtained, with sample-specific abundance values measured for each metabolite.

(2) MetaboLLM generated retrieval-grounded descriptions capturing the biological roles, reaction contexts, and pathway contexts of the measured metabolites. (3) The descriptions were transformed into text embeddings, compared using cosine similarity, and used to select top-ranked metabolite pairs as graph edges. (4) The resulting knowledge-informed topology was shared across samples, while sample-specific metabolite abundances were assigned as node features. (5) A graph isomorphism network (GIN) learned graph-level representations, which were combined with covariates for patient-level prediction of stress hyperglycemia after coronary artery bypass grafting and postmenopausal hormone-regimen classification. (6) Important edges and metabolites were assessed through shared reaction and pathway relationships, randomized graph testing, pathway enrichment, and biological interpretation.

**Stress hyperglycemia prediction**

We evaluated MetaboLLM-GIN in a cohort of 61 patients without pre-existing diabetes who underwent coronary artery bypass grafting (CABG) surgery. Stress hyperglycemia was identified in 36 patients based on perioperative point-of-care glucose measurements collected 4–28 hours after surgery, and 25 did not develop stress hyperglycemia. Metabolomic profiles were obtained at baseline and two hours after surgery using high-resolution liquid chromatography-mass spectrometry. Among the metabolite features annotated using Mummichog[53], 1,100 could be matched to entries in the unified MetaboLLM knowledge database and were retained for downstream analysis.

**Predictive performance for stress hyperglycemia**

Predictive performance was evaluated using five-fold cross-validation repeated across 20 random seeds. The evaluated models included logistic regression (LR), support vector machine (SVM), random forest (RF), extreme gradient boosting (XGBoost), multilayer perceptron (MLP), Feature Tokenizer Transformer (FT-Transformer), and the proposed MetaboLLM-GIN. Conventional machine learning models used features selected through recursive feature elimination followed by partial least squares transformation, whereas MLP and FT-Transformer used the full metabolite feature set. MetaboLLM-GIN represented the 1,100 metabolites as nodes in the LLM-derived graph and used the resulting graph representations for outcome prediction. Statistical significance was assessed using paired t-tests on seed-matched AUC values.

MetaboLLM-GIN achieved the highest mean AUC among all evaluated models at $0.8616 \pm 0.0297$ (Fig. 3a and Table 6), along with the highest accuracy, sensitivity, and specificity, reaching 0.8189, 0.8139, and 0.8260, respectively. Its AUC was significantly higher than that of every competing model across the 20 seeds (all $p < 0.0001$).

**Sensitivity to graph sparsity**

We further evaluated the effect of graph sparsity by varying the number of edges from 100 to 500 (Fig. 3b). The LLM-derived graph achieved its highest AUC at 200 edges and consistently outperformed all alternative graph construction strategies across the full range of graph sizes. Performance declined gradually beyond 200 edges, suggesting that denser graphs may introduce less informative connections that dilute the most relevant biochemical relationships. Based on this analysis, 200 edges were used for graph construction comparisons and subsequent analyses.

**Comparison of graph construction strategies**

To isolate the contribution of graph construction, we trained the same GIN architecture using random, correlation-based, pathway-based, reaction-based, and LLM-derived graphs (Fig. 3a). The density-matched graphs contained 200 edges. We additionally evaluated full curated pathway- and reaction-based graphs that retained all available candidate edges, comprising 29,251 and 2,127 edges, respectively. For the LLM-derived graphs, we compared four variants constructed from the unadapted Qwen backbone with and without retrieval-augmented generation (RAG) and from MetaboLLM with and without RAG.

Among all evaluated graph constructions, the MetaboLLM-derived graph achieved the highest AUC at 0.8616 ± 0.0297, together with the highest accuracy, sensitivity, and specificity (Fig. 3a and Table 6). It outperformed all predictive baselines, conventional graph baselines, and other LLM-derived graph variants in AUC (all $p < 0.0001$). These results indicate that the strongest performance emerged from combining metabolomics-specific adaptation with retrieval-grounded description generation, rather than from the GIN architecture, graph density, or the use of an LLM-generated graph alone.

**Table 6.** Predictive performance and graph construction comparisons for stress hyperglycemia

| Comparison | Method | AUC | Accuracy | Sensitivity | Specificity | p-value |
|---|---|---|---|---|---|---|
| Predictive baselines | LR | 0.8105 ± 0.0327 | 0.7623 ± 0.0279 | 0.7375 ± 0.0498 | 0.7980 ± 0.0740 | <0.0001 |
| | SVM | 0.7556 ± 0.0403 | 0.7147 ± 0.0362 | 0.7333 ± 0.1045 | 0.6880 ± 0.0895 | <0.0001 |
| | RF | 0.7459 ± 0.0501 | 0.7344 ± 0.0378 | 0.7625 ± 0.0751 | 0.6940 ± 0.0864 | <0.0001 |
| | XGBoost | 0.7005 ± 0.0461 | 0.6656 ± 0.0440 | 0.6194 ± 0.0984 | 0.7320 ± 0.1014 | <0.0001 |
| | MLP | 0.7933 ± 0.0295 | 0.7451 ± 0.0274 | 0.7042 ± 0.0535 | 0.8040 ± 0.0634 | <0.0001 |
| | FT-Transformer | 0.6619 ± 0.0342 | 0.6525 ± 0.0339 | 0.6014 ± 0.0955 | 0.7260 ± 0.1212 | <0.0001 |
| Graph baselines | Random | 0.7204 ± 0.0367 | 0.7180 ± 0.0304 | 0.7097 ± 0.0653 | 0.7300 ± 0.0647 | <0.0001 |
| | Correlation | 0.7398 ± 0.0397 | 0.7320 ± 0.0366 | 0.7361 ± 0.0608 | 0.7260 ± 0.0854 | <0.0001 |
| | Pathway | 0.7528 ± 0.0360 | 0.7410 ± 0.0347 | 0.7319 ± 0.0670 | 0.7540 ± 0.0555 | <0.0001 |
| | Pathway (Full) | 0.6203 ± 0.0714 | 0.6410 ± 0.0622 | 0.6125 ± 0.1064 | 0.6820 ± 0.0894 | <0.0001 |
| | Reaction | 0.7602 ± 0.0397 | 0.7402 ± 0.0388 | 0.7347 ± 0.0647 | 0.7480 ± 0.0769 | <0.0001 |
| | Reaction (Full) | 0.7155 ± 0.0438 | 0.7098 ± 0.0455 | 0.7556 ± 0.0784 | 0.6440 ± 0.0788 | <0.0001 |
| LLM-derived graph variants | Qwen3-4B w/o RAG | 0.7931 ± 0.0260 | 0.7533 ± 0.0202 | 0.7236 ± 0.0355 | 0.7960 ± 0.0564 | <0.0001 |
| | Qwen3-4B | 0.7818 ± 0.0279 | 0.7516 ± 0.0257 | 0.7486 ± 0.0427 | 0.7560 ± 0.0710 | <0.0001 |
| | MetaboLLM w/o RAG | 0.7954 ± 0.0309 | 0.7689 ± 0.0344 | 0.7542 ± 0.0600 | 0.7900 ± 0.0755 | <0.0001 |
| | MetaboLLM | **0.8616 ± 0.0297** | **0.8189 ± 0.0294** | **0.8139 ± 0.0534** | **0.8260 ± 0.0665** | |

Best per column in bold.

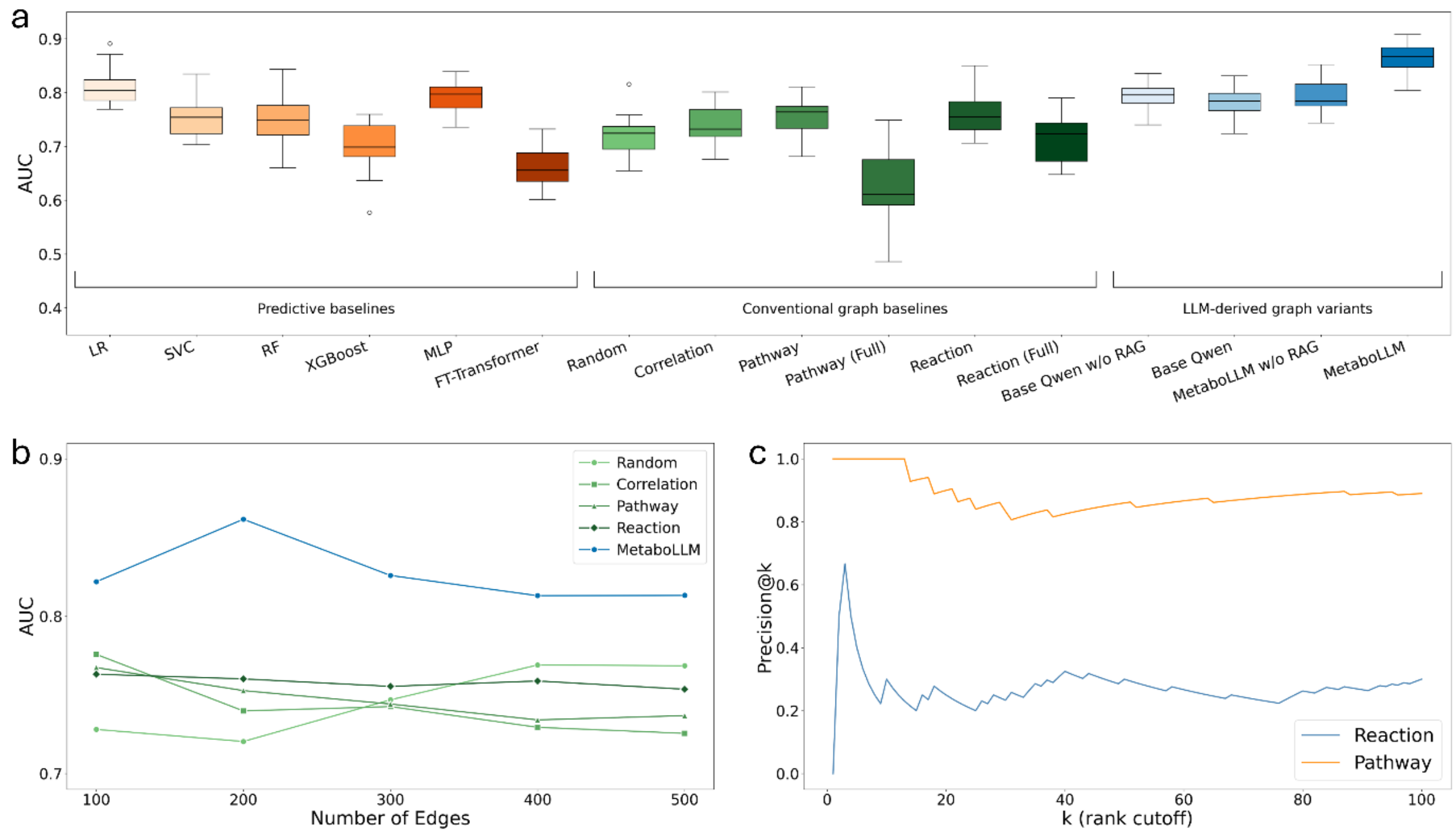

**Fig. 3 | Predictive performance, graph sensitivity, and biological validation for stress hyperglycemia. a** AUC distributions across 20 random seeds for predictive baselines, conventional graph baselines, and LLM-derived graph variants. **b** Mean AUC across graph construction strategies as the number of edges varied from 100 to 500. **c** Precision@k for edges ranked by importance, evaluated according to shared reaction and pathway annotations.

### Biological validation of important graph edges

We first assessed whether edges assigned high importance by GNNExplainer[54] corresponded to known biochemical relationships. Precision@k remained consistently high for shared-pathway relationships, ranging from 0.82 to 1.0 across the evaluated rank cutoffs (Fig. 3c). By contrast, precision for shared reactions was lower and stabilized at approximately 0.3–0.4 for larger values of k. This difference is expected because pathway membership captures broader functional relationships, whereas reaction matching requires both metabolites to participate in at least one common biochemical reaction.

Among the top 100 important edges, 30 connected metabolites participating in a shared reaction and 89 connected metabolites belonging to a shared pathway. Randomized graphs preserving the same node set and number of edges produced mean match ratios of 0.0068 for reactions and 0.0640 for pathways. Both observed ratios were significantly higher than their corresponding null distributions ($p < 0.0001$), indicating that the influential edges connected metabolites related through established biochemical processes rather than arbitrary node pairs.

**Pathway enrichment of important metabolites**

We next examined pathway enrichment among the 100 metabolites assigned the highest node-importance scores. The strongest enrichments were observed for primary bile acid biosynthesis, which included 11 of 39 represented metabolites, and a group of congenital bile acid synthesis disorders, which included 10 of 34 represented metabolites (both $p = 0.0009$). Additional enriched pathways included Vitamin A Deficiency ($p = 0.0010$), retinol metabolism ($p = 0.0012$), and metabolism of xenobiotics by cytochrome P450 ($p = 0.0038$). Detailed overlap counts, pathway sizes, and empirical p-values are provided in Supplementary Table 3.

Lower-ranked enrichments included pathways related to adrenal steroid biosynthesis disorders, steroid hormone biosynthesis, thiopurine drug action, and ubiquinone biosynthesis. Overall, the important metabolites were concentrated in coherent processes involving bile acid, retinoid, steroid-related, and xenobiotic metabolism. Detailed biological interpretation of these pathways and their potential relevance to perioperative stress hyperglycemia is provided in the Discussion.

**Classification of postmenopausal hormone regimens**

We further evaluated MetaboLLM-GIN using the public MTBLS136 metabolomics dataset[55], which included 669 postmenopausal women receiving either estrogen alone ($n = 332$) or estrogen plus progestin ($n = 337$). Unlike the longitudinal measurements used for stress hyperglycemia prediction, this analysis used metabolite abundances measured at a single time point. Among the annotated metabolites, 537 could be matched to entries in the unified MetaboLLM knowledge database and were retained for downstream analysis.

The data were randomly partitioned into 80% training and 20% testing sets across 20 random seeds, and performance was evaluated on the held-out test set for each seed. The same machine learning and deep learning competing models were evaluated, and MetaboLLM-GIN achieved the highest mean AUC at 0.8123 ± 0.0318 (Table 7). It also outperformed all predictive baselines in accuracy, sensitivity, and specificity, reaching 0.7627, 0.7456, and 0.7803, respectively. Its AUC was significantly higher than that of every predictive baseline across the 20 seeds (all $p \leq 0.0085$).

We next compared graph construction strategies after a sensitivity analysis identified 100 edges as the best-performing graph size for this dataset. The density-matched graphs therefore contained 100 edges, while the full pathway- and reaction-based graphs retained all 5,282 and 501 available edges, respectively. We also evaluated LLM-derived graphs generated using the base Qwen model and MetaboLLM, each with and without retrieval augmentation. The final MetaboLLM-derived graph achieved the highest AUC, accuracy, and specificity among all graph constructions and LLM-derived variants (Table 7), although MetaboLLM without RAG showed marginally higher sensitivity. It significantly outperformed all alternative graph constructions and LLM-derived variants in AUC (all $p \leq 0.0152$). These findings reproduce the advantage of the MetaboLLM-derived graph in a larger public dataset under a distinct and non-longitudinal metabolomics classification setting.

**Table 7.** Predictive performance and graph construction for postmenopausal hormone regimen classification

| Comparison | Method | AUC | Accuracy | Sensitivity | Specificity | p-value |
|---|---|---|---|---|---|---|
| Predictive baselines | LR | 0.7781 ± 0.0397 | 0.7272 ± 0.0389 | 0.7235 ± 0.0471 | 0.7311 ± 0.0748 | <0.0001 |
| | SVM | 0.7375 ± 0.1226 | 0.7108 ± 0.0855 | 0.7243 ± 0.0636 | 0.6970 ± 0.1593 | 0.0085 |
| | RF | 0.7610 ± 0.0450 | 0.7187 ± 0.0403 | 0.7096 ± 0.0527 | 0.7280 ± 0.0736 | <0.0001 |
| | XGBoost | 0.7635 ± 0.0449 | 0.7157 ± 0.0385 | 0.7015 ± 0.0653 | 0.7303 ± 0.0705 | <0.0001 |
| | MLP | 0.7944 ± 0.0394 | 0.7463 ± 0.0348 | 0.7382 ± 0.0589 | 0.7545 ± 0.0519 | <0.0001 |
| | FT-Transformer | 0.7133 ± 0.0435 | 0.6780 ± 0.0421 | 0.6809 ± 0.0553 | 0.6750 ± 0.0698 | <0.0001 |
| Graph baselines | Random | 0.7831 ± 0.0404 | 0.7399 ± 0.0357 | 0.7338 ± 0.0829 | 0.7462 ± 0.0819 | <0.0001 |
| | Correlation | 0.8019 ± 0.0370 | 0.7534 ± 0.0318 | 0.7272 ± 0.0539 | 0.7803 ± 0.0424 | 0.0152 |
| | Pathway | 0.7896 ± 0.0353 | 0.7470 ± 0.0369 | 0.7426 ± 0.0686 | 0.7515 ± 0.0500 | 0.0002 |
| | Pathway (Full) | 0.7871 ± 0.0407 | 0.7392 ± 0.0346 | 0.7353 ± 0.0567 | 0.7432 ± 0.0625 | <0.0001 |
| | Reaction | 0.7925 ± 0.0333 | 0.7530 ± 0.0347 | 0.7375 ± 0.0441 | 0.7689 ± 0.0577 | <0.0001 |
| | Reaction (Full) | 0.7961 ± 0.0354 | 0.7418 ± 0.0336 | 0.7426 ± 0.0769 | 0.7409 ± 0.0600 | 0.0104 |
| LLM-derived | Qwen3-4B w/o RAG | 0.7964 ± 0.0344 | 0.7515 ± 0.0340 | 0.7426 ± 0.0535 | 0.7606 ± 0.0653 | <0.0001 |

| graph variants | Qwen3-4B | 0.8060 ± 0.0309 | 0.7563 ± 0.0323 | 0.7478 ± 0.0403 | 0.7652 ± 0.0591 | 0.0132 |
|---|---|---|---|---|---|---|
| | MetaboLLM w/o RAG | 0.8042 ± 0.0326 | 0.7489 ± 0.0305 | **0.7485 ± 0.0540** | 0.7492 ± 0.0495 | 0.0051 |
| | MetaboLLM | **0.8123 ± 0.0318** | **0.7627 ± 0.0302** | 0.7456 ± 0.0432 | **0.7803 ± 0.0413** | |

Best per column in bold.

**Biological validation and pathway interpretation**

Important edges showed significantly greater agreement with known biochemical relationships than expected by chance. Reaction-level evaluation identified two matches among six assessable edges, producing a match ratio of 0.3333 compared with a randomized-graph mean of 0.0433 ($p = 0.0142$). At the pathway level, 14 of 99 evaluated edges connected metabolites sharing at least one pathway, compared with a null mean of 0.0366 ($p < 0.0001$). These findings support the biological relevance of the important edges, although reaction-level evaluation was based on a limited number of annotated edges.

Pathway enrichment among the 100 metabolites ranked highest by node importance identified several significant biological processes. The strongest enrichments were observed for ABC transporters, which included 15 of 44 represented metabolites, and biosynthesis of various other secondary metabolites, which included 6 of 12 represented metabolites ($p = 0.0074$ and $p = 0.0122$, respectively). Additional enriched pathways included amino sugar and nucleotide sugar metabolism ($p = 0.0127$), phenylalanine, tyrosine and tryptophan biosynthesis ($p = 0.0271$), a group of purine-associated disorders ($p = 0.0283$), and insulin resistance ($p = 0.0350$). Detailed overlap counts, pathway sizes, and empirical p-values are provided in Supplementary Table 4.

## Discussion

This study developed MetaboLLM, a metabolomics-specialized language model, and MetaboLLM-GIN, a framework that converts language-encoded biochemical knowledge into graph representations for patient-level metabolomics analysis. The framework connects cross-database knowledge integration, domain adaptation, biochemical description generation, and graph-based prediction. MetaboLLM improved metabolomics knowledge and description generation across multiple backbones, while its retrieval-grounded graph outperformed conventional graph alternatives and other LLM-derived variants in two distinct applications. These findings position a specialized language model not only as a generator of biochemical text but also as a means of organizing heterogeneous knowledge into predictive representations.

This capability differs from existing metabolomics-oriented language systems, which have primarily focused on molecular structure discovery, multimodal metabolite characterization, natural language access to knowledge graphs, or benchmarking existing models rather than adapting a model to integrated relationships among metabolites, pathways, reactions, and enzymes[27–30]. General medical models did not achieve comparable performance, suggesting that adaptation to broad clinical or biomedical text does not necessarily provide the entity-level and relational knowledge required for metabolomics. Improvements across four backbone families further indicate that the benefit was not limited to one model architecture.

The most relevant language model gains occurred in biochemical description generation and relational knowledge, which directly supported downstream graph construction. The staged ablation suggested complementary roles for the two adaptation phases: CPT yielded intermediate gains on internal knowledge tasks and task-dependent improvements in description generation, although it did not improve MetaBench multiple-choice recall, whereas SFT produced consistent gains across both knowledge and generation tasks. This pattern is consistent with CPT enriching domain representations and SFT improving their task-aligned expression. Reaction- and pathway-level retrieval further improved generation without replacing the knowledge acquired during model adaptation. This advantage transferred to the MetaBench Understanding task constructed from PathBank[56], a database not included in the MetaboLLM corpus, supporting the transferability of the learned biochemical representations. Exact symbolic mappings, such as metabolite names and molecular formulas, remained more difficult, but these capabilities were less central to graph construction than the integration of biological roles, pathway context, and biochemical relationships. As supervised training and internal evaluation used GPT-5-mini-generated descriptions, the internal metrics partly measure agreement with teacher references rather than biochemical correctness. The external transfer makes it less likely that the gains were confined to that reference procedure.

The MetaboLLM-derived topology predicted better than graphs built directly from curated pathways or reactions. As descriptions integrate chemical properties, pathway involvement, reaction context, biological roles, and physiological relevance, semantic proximity captures functional relationships that direct curated links do not. This interpretation explains why important edges agreed more strongly with shared pathways than with direct reactions. For example, metabolites may be linked through one or more intermediate metabolites that are absent from the measured dataset. Although a direct reaction graph would leave them disconnected, semantic information may still capture their relationship when they share a strong pathway-level or functional context.

The LLM-variant ablations further showed that this advantage was not explained simply by constructing semantic graphs from descriptions generated by any LLM. Across both applications, graphs derived from retrieval-grounded MetaboLLM descriptions achieved the highest AUC, whereas replacing MetaboLLM with the unadapted Qwen backbone or removing retrieval reduced overall predictive performance. Although the individual effect of retrieval was not uniform across all variants and metrics, these results suggest that metabolomics-specific adaptation and retrieved biochemical context provided the greatest benefit when combined.

The finding that many influential MetaboLLM edges corresponded to known pathways or reactions, while directly constructed curated graphs performed worse, suggests that selecting and prioritizing relationships is as important as determining whether any relationship is documented. The resulting graph is best viewed as a selectively organized functional topology that retains established biochemical structure while connecting indirect or incompletely represented relationships. Its significant enrichment for known pathway and reaction associations supports biological coherence, although semantic edges should not be interpreted as physical interactions or newly validated mechanisms.

Sparse graphs performed best, and performance declined as graph density increased. Adding edges likely introduced weaker or nonspecific semantic relationships that diluted informative signals during message passing. The optimal edge number is unlikely to be universal and may vary with metabolite coverage, annotation availability, and prediction task. As edge number was selected using predictive performance within the same cohorts and the small CABG sample precluded nested selection, absolute performance at the selected graph size is likely optimistic. The MetaboLLM graph remained superior to alternative strategies across the evaluated edge range, indicating that its relative advantage was not confined to one sparsity setting.

The two clinical applications further suggested that graph utility may depend on the information contained in node features. In the CABG cohort, baseline and early postoperative measurements captured both preoperative susceptibility and the acute metabolic response to surgery. Biochemically related metabolites may undergo coordinated perioperative changes, allowing graph propagation to integrate response patterns that are difficult to capture from isolated features. The larger advantage observed in this cohort is therefore consistent with a potential benefit for time-resolved metabolic perturbations. The postmenopausal hormone regimen analysis was not an external validation of hyperglycemia prediction, but it showed that the relative advantage of the LLM-derived graph was reproduced in a larger public dataset using single-time-point measurements.

The metabolites prioritized for stress hyperglycemia prediction mapped to several metabolic processes that have been independently observed in comparable cardiac surgical settings. The enrichment of steroid hormone biosynthesis and related adrenal pathways was consistent with the perioperative endocrine stress response reported during CABG, characterized by concurrent hyperglycemia, elevations in cortisol and catecholamines, and sustained postoperative cortisol elevation[57–59]. Longitudinal metabolomic profiling after on-pump CABG has also demonstrated increased cortisol and decreased dehydroepiandrosterone sulfate, supporting perturbation of the perioperative adrenal response[60]. Primary bile acid biosynthesis pathway was concordant with studies reporting broad postoperative changes in circulating bile acids and associations between bile-acid profiles and outcomes after CABG[60,61]. Retinol-related enrichment similarly agrees with serial measurements showing reduced plasma retinol after CABG and with clinical evidence connecting retinol-binding protein 4 to insulin resistance[62,63]. As enrichment does not establish the direction of pathway activity, these results support pathway perturbation and should remain hypothesis-generating.

The hormone-regimen analysis provided a second, more exploratory interpretation. Enrichment of insulin resistance was consistent with clinical studies showing that hormonal therapy alters glucose and insulin handling and that progestin may attenuate estrogen-associated improvements in insulin sensitivity[64,65]. Aromatic amino-acid and tryptophan-related metabolism is biologically plausible, as estradiol treatment may reduce circulating kynurenic acid and N-acetylkynurenine[64]. In addition, the purine- and urate-related annotations were concordant with studies associating hormone therapy with lower serum uric acid and suggesting that this effect may differ between estrogen-only and estrogen-plus-progestogen therapy[66,67]. Overall, these findings suggest that the important metabolites capture plausible differences in glucose-insulin regulation, tryptophan metabolism, and purine or urate handling between hormone regimens, while remaining exploratory and dependent on the specific metabolites underlying each pathway annotation.

Several limitations should be considered. As internal training and evaluation samples were derived from overlapping public biochemical resources, internal performance does not establish generalization to completely unseen knowledge sources. Transfer to the PathBank-derived MetaBench[30] task partly addresses this concern but cannot exclude overlap among public resources or model pretraining corpora. The evaluation also focused on open general-purpose and medically adapted models, and proprietary frontier LLMs were not directly tested under identical prompting and decoding settings. Although MetaBench provides contextual comparisons with previously reported frontier model results, those comparisons were not fully matched. Accordingly, the results establish the benefit of metabolomics-specific adaptation over the evaluated backbones and medical models rather than superiority over all contemporary LLMs.

Untargeted metabolite identities were based on putative Mummichog annotations[53], and uncertainty in compound matching can propagate through description generation, graph construction, and pathway analysis. Restricting analysis to metabolites matched to the unified database may also favor better-annotated compounds. The primary clinical evaluation involved a small, single-center CABG cohort, and repeated cross-validation does not replace validation in an independent patient population. GNNExplainer scores indicate importance to model predictions rather than causal biological effects, so the identified edges and pathways require experimental or independent clinical confirmation.

The current corpus and prompts primarily encode general biochemical relationships rather than knowledge specific to a disease, treatment, tissue, or objective. This keeps the graph blind to outcomes but may omit relationships that matter only in a given clinical context. Future work could expand model training using curated evidence connecting metabolites to diseases, treatments, tissues, and physiological states, and then condition descriptions on the relevant context while maintaining strict separation from test outcomes. Multi-center and prospective validation will be needed to assess clinical utility. The existing metabolite-reaction-enzyme structure also provides a natural basis for extending the framework toward metabolite-protein-gene graphs and broader multi-omics prediction.

The unified corpus, evaluation benchmark, adapted models, and graph-generation pipeline will be released to support reproducibility and further development. Overall, the findings show that a metabolomics-specialized language model can organize heterogeneous and incomplete biochemical annotations into a ranked functional topology that improves patient-level prediction beyond directly constructed curated graphs. MetaboLLM-GIN provides an initial framework connecting domain-specific language modeling, knowledge-informed graph construction, and biologically interpretable clinical metabolomics analysis.

# Methods

### Knowledge source integration

The metabolomics knowledge corpus was constructed by integrating records from KEGG, HMDB, PubChem, and SMPDB. Metabolite records were aligned across databases using multiple identifier types, including KEGG IDs, HMDB accession numbers, PubChem compound identifiers, InChI keys, and CAS registry numbers. We used a union-find procedure to identify connected components of records that shared at least one identifier, treating each connected component as a unified metabolite entry.

Within each component, source-specific annotations were retained and merged into a single record. The unified entries included compound identities, molecular formulas, chemical classifications, structural identifiers, textual descriptions, pathway memberships, reaction participation, and enzyme annotations when available. Pathway records were aligned between KEGG and SMPDB using available cross-database mappings, and reaction–enzyme relationships were linked through KEGG reaction identifiers and EC numbers.

**Pretraining text generation**

Integrated database records were converted into natural language samples for CPT using template-based generation. We generated compound-, pathway-, and reaction-level text samples rather than training directly on raw structured records, allowing the model to learn metabolomics knowledge in a format consistent with causal language modeling.

Compound-level samples incorporated metabolite names, molecular formulas, chemical classification labels, and available descriptions from HMDB and PubChem. Pathway-level samples used pathway names and descriptions from KEGG and SMPDB. Reaction-level samples used KEGG reaction definitions and included reaction classification and enzyme annotations when available. The predefined templates used for each sample type are provided in Supplementary Information A.2.

**Continual pretraining for metabolomics knowledge**

Qwen3-4B-Instruct was adapted to the metabolomics domain through CPT on the unified metabolomics knowledge corpus. The model was trained using a causal language modeling objective where each token is predicted from its preceding context. This objective enables the model to internalize metabolomics terminology, compound properties, pathway descriptions, reaction definitions, and enzyme-related biochemical context from natural language corpus samples.

CPT was performed using parameter-efficient fine-tuning with low-rank adaptation (LoRA)[68], rather than full model fine-tuning. LoRA freezes the original model weights and introduces trainable low-rank update matrices into selected linear transformations, substantially reducing the number of trainable parameters while preserving capacity for domain adaptation. LoRA was applied to attention projection layers and feedforward network projections.

The CPT stage used compound-, pathway-, and reaction-level samples generated from the integrated metabolomics corpus to enrich the base model with metabolomics-specific factual and relational knowledge before supervised task alignment. Detailed CPT hyperparameters, including LoRA rank, scaling factor, learning rate, batch size, sequence length, optimizer, and scheduler settings, are provided in Supplementary Table 5.

**Supervised fine-tuning for task alignment**

Following CPT, SFT was performed to align the model with metabolomics-specific instruction formats. The CPT checkpoint was used for initialization, allowing the SFT stage to build on the domain knowledge acquired during CPT. SFT used the same LoRA-based parameter-efficient adaptation strategy as CPT and was warm-started from the CPT checkpoint.

SFT examples were formatted as multi-turn conversations containing system, user, and assistant messages. The model was trained using masked causal language modeling, where optimization was applied only to assistant response tokens. This objective encourages the model to generate task-appropriate answers while avoiding optimization on the prompt text itself. Special chat-template tokens were used to distinguish system, user, and assistant turns.

The SFT dataset included short-answer, multiple-choice, and long-answer instruction formats corresponding to the task categories summarized in Table 1. Multiple-choice tasks used task-specific distractor construction to create plausible alternatives while avoiding ambiguous options. Long-answer tasks used curated biochemical reference descriptions, with retrieval-augmented variants constructed for metabolites with available reaction- and pathway-level context. Detailed procedures for sample filtering, distractor generation, prompt variation, answer formatting, train/test splitting, and SFT hyperparameters are provided in Supplementary Information B and Supplementary Table 6.

**Retrieval-augmented description generation**

Retrieval-augmented description generation was used to construct grounded metabolite-description examples for structure-rich compounds. For each target metabolite, context retrieval followed an identifier-constrained pipeline. The target metabolite was first matched in the vector database using its primary available identifier, prioritized as KEGG ID, HMDB ID, and then PubChem ID. Reactions containing the target metabolite were retrieved using matched metabolite identifiers, and the metabolite identifiers in those reaction records were used as metadata filters to define reaction-linked candidate sets. Pathways containing the target metabolite were retrieved using the same identifier-based strategy, and pathway-linked metabolites were used to provide broader biochemical context.

Candidate metabolites were ranked using complementary biological and chemical similarity scores. ChemBERTa[69] embeddings were used to compute chemical similarity from SMILES representations[70], whereas PubMedBERT[71] embeddings were used to compute biological similarity from textual metabolite descriptions. The two similarity scores were averaged using equal weights to produce the final ranking score.

Retrieved reaction- and pathway-level contexts were incorporated into structured prompts for structure-rich metabolite description generation. Structure-poor metabolite descriptions followed the same general instruction format but did not include external reaction or pathway context. Responses for both regimes were generated using GPT-5-mini[72] and appended to the conversational SFT samples. Target metabolites in the structure-rich regime were stratified by annotation richness to ensure coverage across different levels of available biochemical information. Detailed retrieval settings, including identifier priority, retrieved candidate numbers, similarity weights, redundancy filtering, and prompt construction parameters, are provided in Supplementary Table 7 and Supplementary Information B.3.

**LLM-based graph construction**

A key component of MetaboLLM-GIN is the construction of metabolite graphs from the biochemical knowledge encoded by the trained metabolomics language model. Rather than defining graph connectivity solely from incomplete pathway annotations or fixed correlation thresholds, we used MetaboLLM-generated metabolite descriptions to derive edges that reflect semantic similarity in biochemical function, pathway involvement, and metabolic context. The resulting graph provides a shared knowledge-informed structure for downstream metabolomics prediction.

For a downstream dataset containing metabolites $\mathcal{M} = \{m_1, m_2, \ldots, m_n\}$, we first generated a textual representation for each metabolite using MetaboLLM with retrieval augmentation. For each metabolite $m_i$, a query was constructed and relevant biochemical context was retrieved using the RAG pipeline described in the previous section and Supplementary Information B.3. The model then generated a response $r_i$ summarizing the metabolite's biological role, pathway associations, reaction-level context, and metabolic significance. These responses served as language-based representations of metabolites and provided the basis for graph construction.

To convert the generated descriptions into graph edges, we embedded each response using the same biomedical text embedding model used in the retrieval system. For each description $r_i$, an embedding vector $\mathrm{e}_i \in \mathbb{R}^d$ was obtained. We then computed pairwise cosine similarity between all metabolite embeddings to construct a metabolite similarity matrix:

$$S_{ij} = \frac{\mathbf{e}_i \cdot \mathbf{e}_j}{\|\mathbf{e}_i\| \|\mathbf{e}_j\|}.$$

The diagonal entries of $S$ were set to zero to exclude self-loops during edge selection. To define the adjacency structure, we selected the top-$k$ metabolite pairs with the highest similarity scores from the upper triangular portion of $S$. These selected pairs were treated as undirected edges and represented bidirectionally for GIN input. The edge number $k$ was treated as a graph sparsity parameter.

For each sample $s$, we constructed a graph $G_s = (V, E, X_s)$, where $V = \mathcal{M}$ denotes the metabolite node set, $E$ denotes the shared MetaboLLM-derived edge set, and $X_s \in \mathbb{R}^{n \times q}$ denotes the sample-specific node feature matrix. The graph topology is fixed across samples within a dataset because $E$ is inferred from metabolite-level biochemical descriptions, whereas $X_s$varies according to measured metabolite profiles. The feature dimension $q$ depends on the available input measurements and may include a single metabolite intensity, multiple observed time points, or derived temporal features such as changes between clinically relevant time points. This formulation allows the same knowledge-informed graph structure to support both static and time-resolved metabolomics prediction tasks.

**Alternative and ablation graph construction strategies**

To compare the proposed topology against alternative graph definitions, we evaluated both LLM-derived ablations and conventional graph construction strategies using the same metabolite node set. To assess the contributions of metabolomics-specific adaptation and retrieved biochemical context, the proposed retrieval-grounded MetaboLLM topology was compared against three LLM-derived ablations: Qwen3-4B without retrieval, Qwen3-4B using retrieval, and MetaboLLM without retrieval. All four LLM-derived configurations used the same description embedding, cosine-similarity calculation, top-k edge selection, GIN architecture, and training procedure, differing only in the language model and the inclusion of retrieved context.

We additionally constructed reaction-based, pathway-based, correlation-based, random, and full curated graphs. Reaction-based candidate edges connected metabolite pairs participating in at least one common KEGG reaction, whereas pathway-based candidate edges connected pairs sharing at least one KEGG or SMPDB pathway. When either candidate set exceeded the target graph size $k$, $k$ edges were sampled uniformly and independently for each random seed. Correlation-based graphs retained the k metabolite pairs having the highest absolute Spearman rank correlations, which were used to reduce sensitivity to skewed abundance distributions and tied zero values arising from measurements below the detection limit. Random graphs sampled k pairs uniformly from all possible node pairs for each seed. Full curated graphs retained all available reaction- or pathway-based candidate edges without imposing a density constraint.

All selected pairs were treated as undirected edges and represented bidirectionally for GIN input. The target graph size was set to $k$=200 for the stress hyperglycemia cohort based on the edge-number sensitivity analysis and to $k$=100 for postmenopausal hormone regimen classification. Node features, model architecture, and training procedures remained unchanged across graph construction strategies.

**Graph isomorphism network-based classification**

We used a graph isomorphism network (GIN) for graph-level classification on the graphs constructed above. GIN updates each metabolite representation by combining its own features with aggregated information from neighboring metabolites, allowing the model to capture local biochemical context defined by the LLM-derived graph.

Let $H^{(0)} = X_s$ denote the input node feature matrix for sample $s$. At the $l$-th GIN layer, the representation of node $v$ was updated as

$$h_v^{(l)} = MLP^{(l)}\left(\left(1 + \epsilon^{(l)}\right)h_v^{(l-1)} + \sum_{u \in \mathcal{N}(v)} h_u^{(l-1)}\right),$$

where $\mathcal{N}(v)$ denotes the neighbors of node $v$, $\epsilon^{(l)}$ is a learnable parameter, and $\text{MLP}^{(l)}$ denotes the layer-specific multilayer perceptron. Stacking multiple GIN layers enables information propagation beyond directly connected metabolites.

For graph-level prediction, we used readouts from both the original node features and the hidden representations produced by each GIN layer. Each representation level was passed through an MLP readout to obtain a fixed-dimensional graph embedding, and the embeddings were concatenated:

$$z_s = \text{Concat}\left(\rho^{(0)}(H^{(0)}), \rho^{(1)}(H^{(1)}), \ldots, \rho^{(L)}(H^{(L)})\right),$$

where $\rho^{(l)}(\cdot)$ denotes the readout MLP at level $l$. When clinical covariates were available, they were concatenated with $z_s$ before the final classifier. The resulting representation was passed through an MLP classifier to produce a binary outcome logit and trained using binary cross-entropy loss with logits. Model hyperparameters and dataset-specific preprocessing procedures are provided in Supplementary Tables 8 and 9.

**Model interpretation and biological validation**

To interpret predictions from MetaboLLM-GIN, GNNExplainer was applied according to the evaluation design used for each application[54]. Stress hyperglycemia prediction used five-fold cross-validation across 20 random seeds, whereas postmenopausal hormone-regimen classification used 80:20 train-test splits across 20 random seeds. GNNExplainer learned sparse node and edge masks that preserved model predictions, and importance scores were aggregated across samples and seeds to obtain dataset-level rankings.

For node-level interpretation, the top 100 metabolites were mapped to the unified knowledge database using KEGG, HMDB, and PubChem identifiers and assessed for pathway enrichment relative to the full ranked metabolite set. For each pathway, the observed overlap among the top 100 metabolites was compared with overlaps obtained from 10,000 random selections of 100 metabolites from the full metabolite set, and empirical p-values were calculated as the proportion of randomized overlaps equal to or greater than the observed value. For edge-level validation, the top 100 ranked undirected edges were evaluated separately for shared reactions and pathways. An edge was considered a reaction match when both metabolites participated in at least one common KEGG reaction and a pathway match when they shared at least one KEGG or SMPDB pathway. The observed match ratios were compared with null distributions generated from 10,000 randomized graphs containing the same nodes and number of edges, with empirical p-values calculated as the proportion of randomized ratios equal to or greater than the observed ratio.

## Data availability

The MetaboLLM-Corpus is publicly available through the MetaboLLM organization on Hugging Face at https://huggingface.co/datasets/MetaboLLM. MTBLS136 is publicly available through MetaboLights. Stress hyperglycemia data are available from Francisco J. Pasquel upon reasonable request, subject to applicable institutional and ethical approvals.

## Code and model availability

The source code for model adaptation, evaluation, and MetaboLLM-GIN is publicly available at https://github.com/dohyunku9/MetaboLLM. The adapted MetaboLLM model weights are available through the MetaboLLM organization on Hugging Face at https://huggingface.co/MetaboLLM.

## Acknowledgements

We thank M. Ryan Smith (Atlanta Veterans Affairs Health Care System) for methodological guidance on metabolite, reaction, and pathway analyses. Min Gu Kwak's contributions to this work were completed before he joined Capital One.

# Supplementary Information

### A. CPT Data Construction Details

This appendix provides detailed information on the continual pretraining (CPT) data construction process, including database integration procedures, template specifications, and corpus statistics.

### A.1 Database Integration Pipeline

The integration of metabolomics databases proceeds through three stages: individual database preprocessing, cross-database alignment, and record merging.

Individual Database Preprocessing: We preprocess each source database to extract relevant fields and standardize formats. From KEGG, we extract compound identifiers, names, molecular formulas, compound classification, pathway associations, reaction participations, and enzyme annotations. From HMDB, we extract accession numbers (including secondary accessions), compound names, descriptions, chemical formulas, structural representations (SMILES, InChI, InChIKey), and pathway mappings. From PubChem, we extract compound identifiers, names, molecular formulas, structural representations, and descriptive text. SMPDB pathway records are also processed to obtain pathway names, descriptions, and metabolite-pathway mappings.

Cross-Database Alignment: We align records across databases using multiple identifier types to maximize linkage. The alignment employs PubChem's CID-InChIKey mappings to connect HMDB compounds (which contain InChIKeys) to PubChem entries, and CID-SID mappings to link KEGG compounds (which reference PubChem substance IDs) to PubChem compound records. We then apply a union-find algorithm that treats each database record as a node and creates edges between records sharing any common identifier (KEGG ID, PubChem ID, HMDB ID, InChI, InChIKey, or CAS number). Connected components in this graph represent the same underlying compound across databases.

Record Merging: Within each connected component, we merge information following source priority rules. For molecular formulas, we prioritize PubChem (computed), then HMDB, then KEGG. For structural representations, we canonicalize SMILES using RDKit to ensure consistency, prioritizing InChI-derived SMILES, then PubChem SMILES, then HMDB SMILES. Textual descriptions are retained separately by source to enable template-based combination during text generation. Compound classification fields are also retained when available and later inserted into compound-level CPT samples using classification templates. Pathway associations are combined from both KEGG pathway IDs and SMPDB IDs, excluding overly general pathways (map01100: Metabolic pathways, map01110: Biosynthesis of secondary metabolites, map01120: Microbial metabolism in diverse environments) that provide limited specificity. Reaction records and enzyme annotations are retained as separate reaction-level resources and linked through KEGG reaction information and EC numbers for reaction sample generation.

**A.2 Template Specifications**

We generate CPT text using fixed natural language templates that transform structured records, including compound, pathway, reaction with enzymes, into sentences. Templates are grouped by content type, including compound templates with molecular formulas, compound templates without formulas, pathway insertion statements appended to compound samples, compound classification insertion statements, description combination templates, pathway description combination templates, pathway templates, reaction templates, reaction classification insertion statements, and enzyme insertion templates.

Compound templates populate placeholders such as {name_part}, {formula}, and {description} to describe chemical identity and properties. Pathway templates use {pathway_name} and {description} to generate standalone pathway-level text, while pathway insertion templates attach pathway information to compound descriptions. Classification templates use {meta_level_1} and {meta_level_2} to append compound or reaction category information when available. Reaction templates use {definition} to describe biochemical transformations, and enzyme insertion templates use {enzyme_names} and, when available, {enzyme_comments} to add catalytic context. For each sample, a template is selected at random from the corresponding group to introduce variation in phrasing while preserving consistent biochemical content.

Compound Templates with Formula:

"{name_part} has the molecular formula {formula}. {description}"

"{name_part} is a compound with the formula {formula}. {description}"

"The compound {name_part} has a molecular formula of {formula}. {description}"

"The molecular formula of {name_part} is {formula}. {description}"

"With the molecular formula {formula}, {name_part} is known for the following properties. {description}"

Compound Templates without Formula:

"{name_part} is a biochemical compound found in metabolic processes. {description}"

"The compound {name_part} plays a role in biological systems. {description}"

"This compound is commonly referred to as {name_part}. {description}"

"{name_part} is a metabolite involved in various biochemical pathways. {description}"

"As a biochemical compound, {name_part} has been studied extensively. {description}"

Pathway Insertion Templates appended to compound samples:

"This compound is involved in {pathways} pathways."

"It participates in {pathways} pathways."

"It plays a role in {pathways} pathways."

"This metabolite is part of {pathways} pathways."

"It is associated with {pathways} pathways."

Compound Classification Templates appended to compound samples:

Templates when both classification levels are available:

"This compound is classified as {meta_level_1}. Under this class, it is further categorized as {meta_level_2}."

"It belongs to the {meta_level_1} class and is specifically categorized as {meta_level_2}."

"As a member of the {meta_level_1} category, this compound is further classified as {meta_level_2}."

"This metabolite falls under the {meta_level_1} classification, with a more specific category of {meta_level_2}."

"Classified as {meta_level_1}, this compound is subcategorized as {meta_level_2}."

Templates when only the first classification level is available:

"This compound is classified as {meta_level_1}."

"It belongs to the {meta_level_1} class."

"This metabolite is categorized as {meta_level_1}."

"The compound is grouped under {meta_level_1}."

"This compound falls under the {meta_level_1} classification."

Templates when only the second classification level is available:

"This compound belongs to the {meta_level_2} category."

"It is classified as {meta_level_2}."

"This metabolite is categorized as {meta_level_2}."

"This substance is identified as a {meta_level_2}."

"This compound falls under the {meta_level_2} classification."

Description Combination Templates when both HMDB and PubChem descriptions are available:

"{hmdb_desc} According to other sources, {pubchem_desc}"

"{hmdb_desc} It is also described as: {pubchem_desc}"

"{hmdb_desc} Additionally, other sources note that {pubchem_desc}"

Pathway Description Combination Templates when both SMPDB and KEGG descriptions are available:

"{smpdb_desc} According to other sources, {kegg_desc}"

"{smpdb_desc} It is also described as: {kegg_desc}"

"{smpdb_desc} Additionally, other sources note that {kegg_desc}"

Pathway Templates:

"{pathway_name} is a metabolic pathway. {description}"

"The {pathway_name} pathway is a biochemical process. {description}"

"{pathway_name} corresponds to a biochemical pathway. {description}"

"{pathway_name} is a biochemical pathway involved in metabolism. {description}"

"The metabolic pathway {pathway_name} plays an important role in cellular processes. {description}"

Reaction Templates:

"The following reaction occurs in biological systems: {definition}."

"A biochemical reaction is defined as: {definition}."

"This biochemical reaction is defined as follows: {definition}."

"In metabolism, this reaction takes place: {definition}."

"The biochemical transformation is described as: {definition}."

Reaction Classification Templates appended to reaction samples:

Templates when both reaction classification levels are available:

"This reaction is classified as {meta_level_1}. Under this class, it is further categorized as {meta_level_2}."

"It belongs to the {meta_level_1} class and is specifically categorized as {meta_level_2}."

"As a member of the {meta_level_1} category, this reaction is further classified as {meta_level_2}."

"This reaction falls under the {meta_level_1} classification, with a more specific category of {meta_level_2}."

"Classified as {meta_level_1}, this reaction is subcategorized as {meta_level_2}."

Templates when only the first reaction classification level is available:

"This reaction is classified as {meta_level_1}."

"It belongs to the {meta_level_1} class."

"This reaction is categorized as {meta_level_1}."

"The reaction is grouped under {meta_level_1}."

"This reaction falls under the {meta_level_1} classification."

Templates when only the second reaction classification level is available:

"This reaction belongs to the {meta_level_2} category."

"It is classified as {meta_level_2}."

"This reaction is categorized as {meta_level_2}."

"This reaction is identified as a {meta_level_2}."

"This reaction falls under the {meta_level_2} classification."

Enzyme Insertion Templates appended to reaction samples:

Templates when enzyme names and comments are available:

"This reaction is catalyzed by {enzyme_names}. {enzyme_comments}"

"The enzyme(s), {enzyme_names}, catalyze this reaction. {enzyme_comments}"

"{enzyme_names} is responsible for this biochemical transformation. {enzyme_comments}"

"This reaction is facilitated by {enzyme_names}. {enzyme_comments}"

"The catalytic activity is performed by {enzyme_names}. {enzyme_comments}"

Templates when only enzyme names are available:

"This reaction is catalyzed by {enzyme_names}."

"The enzyme(s), {enzyme_names}, catalyze this reaction."

"{enzyme_names} is responsible for this biochemical transformation."

"This reaction is facilitated by {enzyme_names}."

"The catalytic activity is performed by {enzyme_names}."

Enzyme Comment Attribution Templates:

"Regarding {enzyme_name}: {comment}"

"About {enzyme_name}: {comment}"

"As for {enzyme_name}: {comment}"

"Concerning {enzyme_name}: {comment}"

"{enzyme_name} is characterized as follows: {comment}"

### A.3 Text Preprocessing

All generated text is preprocessed to ensure consistency and to remove formatting artifacts. Character normalization is applied by converting smart quotes to standard quotes, normalizing dash variants to hyphens, replacing non-breaking spaces with regular spaces, and removing zero-width characters. Whitespace is standardized by stripping leading and trailing spaces and collapsing multiple consecutive spaces into a single space. Samples with empty descriptions after preprocessing are excluded from the training corpus.

We analyze token length distributions using the target model's tokenizer to determine an appropriate sequence length for training. Based on this analysis, we set the maximum sequence length to 1,536 tokens, which covers approximately the 95th percentile of generated samples while balancing computational efficiency.

## B. SFT Data Construction Details

### B.1 MCQA Distractor Generation

For MCQA tasks, we generated three distractor options using task-specific strategies designed to create plausible but distinguishable alternatives. Formula-based tasks used chemical formula similarity, whereas pathway, class identification, and relation identification tasks used database-derived exclusion rules to avoid ambiguous or trivially correct choices. Across all MCQA tasks, the correct answer and distractors were randomly shuffled before assigning answer labels.

### B.1.1 Formula-Based Distractor Generation

For molecular formula tasks, chemical formulas were parsed into element count dictionaries using regular expression matching. Elements without explicit counts were assigned a count of 1.

Example: "C6H12O6" → {C: 6, H: 12, O: 6}

Example: "NaCl" → {Na: 1, Cl: 1}

For two parsed formulas $p_1$ and $p_2$, formula similarity was computed using a count-based Jaccard similarity:

$$J(p_1, p_2) = \frac{\sum_{e \in E} \min(p_1[e], p_2[e])}{\sum_{e \in E} \max(p_1[e], p_2[e])}$$

where $E$ is the union of all elements in both formulas, and $p[e]$ denotes the count of element $e$, with absent elements assigned a count of 0.

To improve computational efficiency, we constructed a formula similarity index. For datasets with more than 2,000 compounds, each compound was compared against a random sample of 2,000 other compounds. For smaller datasets, full pairwise comparisons were performed. Compounds with identical SMILES representations were excluded from similarity consideration to avoid selecting the same molecule under different names. Candidate compounds with formula similarity greater than 0.3 were retained, sorted in descending order, and the top 20 candidates were stored for each compound.

For Name → Formula MCQA, distractor formulas were selected from the precomputed formula similarity index. Up to 10 highly similar candidates were considered first, candidates with formulas identical to the correct answer were removed, and random compounds were used as fallback candidates when fewer than three unique distractors were available. For Formula → Name MCQA, up to 15 similar candidates were considered, and compounds sharing the same molecular formula as the target compound were excluded because they could also be valid answers. Random compounds were again used as fallback candidates when fewer than three valid distractors were available.

### B.1.2 Pathway-Based Distractor Generation

For pathway-related knowledge tasks, distractors were generated using compound–pathway membership information from the integrated database. In the Metabolite → Pathway task, the correct answer was a pathway associated with the input metabolite, and distractor pathways were selected from pathways not associated with that metabolite. Trivial pairs in which the compound name appeared directly in the pathway name were excluded to reduce string-matching shortcuts.

In the Pathway → Metabolite task, the correct answer was a metabolite belonging to the input pathway. To reduce ambiguity, metabolites appearing in more than 20 pathways were excluded because they represent overly general compounds that are not pathway-specific. Distractor metabolites were selected from compounds that did not share any pathway with the correct answer, ensuring that distractors were pathway-disjoint from the target relationship.

#### B.1.3 Class Identification Distractor Generation

For class identification tasks, distractors were sampled from biochemical class labels or entities outside the correct class, depending on the task direction. Both Level 1 and Level 2 classification labels were used when available. When both levels were present for an entity, one level was selected during data generation.

For Metabolite → Class and Reaction → Class tasks, distractor class labels were sampled from the same classification level as the correct answer. Class labels with substring overlap with the correct class were excluded to reduce ambiguous options caused by nested or highly similar class names. For Class → Metabolite and Class → Reaction tasks, the correct answer was selected from entities belonging to the given class, whereas distractors were selected from entities outside that class. This ensured that each option represented a plausible biochemical entity while maintaining class-level discriminability.

#### B.1.4 Relation Identification Distractor Generation

For relation identification tasks, distractors were generated from database-derived metabolite–reaction and reaction–enzyme associations. In the Metabolite–Reaction Relation task, the correct answer was a metabolite that shared at least one biochemical reaction with the input metabolite. Metabolites involved in more than 20 reactions were excluded to remove highly connected hub-like entities. Distractor metabolites were required to be reaction-disjoint from both the input metabolite and the correct answer, preventing alternative options from also satisfying the shared-reaction relation.

For the Reaction → Enzyme task, the correct answer was an enzyme annotated as catalyzing the given reaction. Distractor enzymes were selected from enzymes not associated with that reaction. When possible, distractor selection considered enzyme class structure to provide plausible alternatives while avoiding enzymes that would also be valid for the target reaction.

For the Reaction–Enzyme Pair Validation task, the correct option was a biologically valid reaction–enzyme pair. Distractor options were constructed from mismatched reaction–enzyme pairs so that the reaction and enzyme were individually valid database entities but did not form the annotated pair.

### B.1.5 MCQA Label Format Variations

To improve robustness to answer format variations, we randomly select from multiple label formats during data generation:

| Format Style | Labels |
|---|---|
| Period | A. B. C. D. |
| Colon | A: B: C: D: |
| Parenthesis-right | A) B) C) D) |
| Parenthesis-both | (A) (B) (C) (D) |
| Lowercase-right | a) b) c) d) |
| Lowercase-both | (a) (b) (c) (d) |
| Lowercase-colon | a: b: c: d: |
| Bracket-upper | [A] [B] [C] [D] |
| Bracket-lower | [a] [b] [c] [d] |

Additionally, option headers are randomly varied: “Options:”, “Choices:”, “Select the correct option:”, “Candidates:”, or no header.

## B.2 Prompt Template Specifications

Each reported SFT task used multiple randomly sampled system and user prompt variants to increase linguistic diversity. System prompt pools generally contained 6–7 variants per task, while user prompt pools generally contained 4–5 variants per task. To keep the appendix concise, two example system prompts and two example user prompts are shown for each reported task. For MCQA tasks, prompts instructed the model to return only the answer label. Detailed RAG and non-RAG prompt construction is provided in Appendix B.3.

### B.2.1 Knowledge Recall Prompt Templates

Knowledge recall prompts covered compound identity and pathway membership tasks. The Name → Formula task was implemented in both short-answer and MCQA formats, whereas Formula → Name, Metabolite → Pathway, and Pathway → Metabolite tasks were implemented in MCQA format.

***Name → Formula, short-answer***

System prompt examples:
"You are a metabolomics expert. Provide only the exact chemical formula. Use standard notation (e.g. C6H12O6). No explanations."
"State the molecular formula. Nothing else. Format: CxHyOz (e.g. C6H12O6)."

User prompt examples:
"What is the chemical formula of {name}?"
"Provide the molecular formula for {name}."

***Name → Formula, MCQA***

System prompt examples:
"You are a metabolomics expert specializing in molecular formulas. Select ONLY from the given options. One of the options is always correct. Output only (A), (B), (C), or (D)."
"As a chemistry tutor, identify the correct molecular formula. One of the options is always correct. Reply with just the letter."

User prompt examples:
"Choose the correct chemical formula for {name} from the options below."
"Which of the following is the molecular formula of {name}?"

***Formula → Name, MCQA***

System prompt examples:
"You are a metabolite nomenclature expert. Which compound name is correct? One of the options is always correct. Reply with only (A), (B), (C), or (D)."
"As a biochemist, identify the correct metabolite name. One option is always correct. Just the letter."

User prompt examples:
"Which compound has the formula {formula}?"
"Select the name of the compound with molecular formula {formula}."

***Metabolite → Pathway, MCQA***

System prompt examples:
"You are a metabolomics expert specializing in pathway analysis. Which pathway is correct? One of the options is always correct. Reply with only (A), (B), (C), or (D)."
"As a pathway biochemist, identify the correct metabolic pathway. One option is always correct. Just the letter."

User prompt examples:
"Which metabolic pathway involves {name}?"
"Select the pathway that includes {name}."

***Pathway → Metabolite, MCQA***

System prompt examples:
"You are a metabolomics expert. Which metabolite belongs to this pathway? One of the options is always correct. Reply with only (A), (B), (C), or (D)."
"As a pathway biochemist, identify the correct compound. One option is always correct. Just the letter."

User prompt examples:
"Which metabolite is part of the {pathway_name}?"
"Select a compound involved in the {pathway_name}."

### B.2.2 Class Identification Prompt Templates

Class identification prompts covered both entity-to-class and class-to-entity directions for metabolites and reactions.

***Metabolite → Class, MCQA***

System prompt examples:
"You are a metabolomics expert specializing in compound classification. Which class is correct? One of the options is always correct. Reply with only (A), (B), (C), or (D)."
"As a biochemistry classifier, identify the correct metabolite class. One option is always correct. Just the letter."

User prompt examples:
"Which metabolite class does {name} belong to?"
"Select the correct classification for the metabolite {name}."

***Class → Metabolite, MCQA***

System prompt examples:
"You are a metabolomics expert. Which metabolite belongs to this class? One of the options is always correct. Reply with only (A), (B), (C), or (D)."

"As a biochemistry classifier, identify the metabolite that belongs to this class. One option is always correct. Just the letter."

User prompt examples:
"Which of the following metabolites belongs to the class {class_name}?"
"Select the metabolite that is classified under {class_name}."

***Reaction → Class, MCQA***

System prompt examples:
"You are a biochemistry expert specializing in reaction classification. Which class is correct? One of the options is always correct. Reply with only (A), (B), (C), or (D)."
"As a reaction classifier, identify the correct reaction class. One option is always correct. Just the letter."

User prompt examples:
"Which reaction class does {reaction_name} belong to?"
"Select the correct classification for the reaction {reaction_name}."

***Class → Reaction, MCQA***

System prompt examples:
"You are a biochemistry expert. Which reaction belongs to this class? One of the options is always correct. Reply with only (A), (B), (C), or (D)."
"As a reaction classifier, identify the reaction that belongs to this class. One option is always correct. Just the letter."

User prompt examples:
"Which of the following reactions belongs to the class {class_name}?"
"Select the reaction that is classified under {class_name}."

### B.2.3 Description Generation Prompt Templates

Description generation prompts were used for standard long-answer metabolite, pathway, and enzyme description tasks.

***Metabolite Description, long-answer***

System prompt examples:
"You are a metabolomics expert. Explain this compound's role in metabolism and its biological significance."
"As a biochemistry professor, describe this metabolite comprehensively. Cover its function and properties."

User prompt examples:
"Explain the role of {name} in human metabolism."
"Describe the biochemical function of {name}."

***Pathway Description, long-answer***

System prompt examples:
"You are a metabolomics expert. Explain the metabolic pathway in detail, including its biological significance."
"As a biochemistry professor, describe this pathway comprehensively. Cover its mechanism and physiological role."

User prompt examples:
"Provide a comprehensive scientific summary of the {pathway_name} pathway, including its biological significance."
"Describe the {pathway_name} pathway and its role in metabolism."

***Enzyme Description, long-answer***

System prompt examples:
"You are a biochemistry expert specializing in enzymology. Explain the enzyme's function, mechanism, and biological significance."
"As an enzymology professor, describe this enzyme comprehensively. Cover its catalytic mechanism and physiological role."

User prompt examples:
"Describe the enzyme {enzyme_name} and its biochemical function."
"Explain the role and mechanism of the enzyme {enzyme_name}."

#### B.2.4 Relation Identification Prompt Templates

Relation identification prompts covered metabolite–reaction and reaction–enzyme associations.

***Metabolite–Reaction Relation, MCQA***

System prompt examples:
"You are a metabolomics expert specializing in reaction networks. Which metabolite shares a reaction? One of the options is always correct. Reply with only (A), (B), (C), or (D)."
"As a biochemistry expert, identify the metabolite that participates in the same reaction. One option is always correct. Just the letter."

User prompt examples:
"Which metabolite participates in the same reaction as {name}?"
"Select the compound that shares a biochemical reaction with {name}."

***Reaction → Enzyme, MCQA***

System prompt examples:
"You are an enzymology expert. Which enzyme catalyzes this reaction? One of the options is always correct. Reply with only (A), (B), (C), or (D)."
"As a biochemistry expert, identify the enzyme that facilitates this reaction. One option is always correct. Just the letter."

User prompt examples:
"Which enzyme catalyzes the following reaction?\n{definition}"
"Identify the enzyme that facilitates this reaction:\n{definition}"

***Reaction–Enzyme Pair Validation, MCQA***

System prompt examples:
"You are a biochemistry expert. Which reaction-enzyme pair is correctly matched? One of the options is always correct. Reply with only (A), (B), (C), or (D)."
"As a metabolic pathway analyst, identify the valid reaction-enzyme combination. One option is always correct. Just the letter."

User prompt examples:
"Which reaction-enzyme pair is correctly matched?"
"Select the correct pairing of reaction and its catalyzing enzyme."

### B.3 RAG Data Pipeline

The RAG dataset construction used a metadata-filtered retrieval pipeline designed to provide structured biochemical context for each target metabolite. Rather than relying solely on global text similarity, the pipeline first identified candidate metabolites through database identifiers and curated biochemical relationships, then ranked candidates using embedding-based similarity.

#### B.3.1 Vector Database Architecture

Metabolomics data were stored in a Qdrant vector database with three collections:

1. Biochemical compounds: metabolite records with text representations, chemical and biological embeddings, and cross-database identifiers, including KEGG ID, HMDB ID, and PubChem ID.

2. Reactions: biochemical reaction records linking metabolites through enzymatic transformations. Each record included reaction text, participating metabolite identifiers, and enzyme annotations.
3. Pathways: metabolic pathway records with pathway descriptions and participating metabolite identifiers.

This database structure enabled metadata-filtered retrieval of biochemically related metabolites before embedding-based ranking.

### B.3.2 Metadata-Filtered Candidate Retrieval

For each target metabolite, context retrieval proceeded through a three-step pipeline.

*Step 1: Input Metabolite Lookup.* Given a target metabolite identifier, we query the biochemical compounds collection using ID priority (KEGG > HMDB > PubChem). This retrieves the metabolite's stored properties, text description, and embeddings.

*Step 2: Reaction-Based Retrieval.* We identify metabolites that directly interact with the input through biochemical reactions:

1. Query the reactions collection for all reactions containing the input metabolite
2. Extract all other metabolite IDs from these reactions as candidates
3. For each candidate, compute weighted similarity to the input: $s_{combined} = w_{bio} \cdot s_{bio} + w_{chem} \cdot s_{chem}$
4. Filter candidates exceeding the duplicate similarity threshold (0.98)
5. Apply name-based deduplication to remove identical metabolites with different IDs
6. Select top-$k$ candidates ranked by similarity
7. Each selected metabolite is annotated with the reaction texts linking it to the input, providing explicit biochemical context.

*Step 3: Pathway-Based Retrieval.* We identify metabolites that co-participate in pathways with the input:

1. Query the pathways collection for all pathways containing the input metabolite
2. If more pathways exist than the limit, rank by biological similarity to the input and select top pathways
3. For each selected pathway:
    a. Extract metabolite IDs, excluding the input and previously selected metabolites
    b. Compute weighted similarity for each candidate
    c. Apply duplicate and name-based filtering
    d. Select top metabolites per pathway

This hierarchical approach ensures the final context includes both direct biochemical reactions and broader pathway context. The default values of these retrieval parameters are summarized in the Supplementary Table 7.

**B.3.3 RAG Prompt Construction**

RAG prompts were constructed using a structured context format that interleaved task instructions with retrieved biochemical knowledge. The user prompt consisted of the following sections:

================================================================================

## Task

[Task description requesting a 300–350 word metabolite description]

## Input Metabolite

Names: [deduplicated names across KEGG, HMDB, PubChem]

Description: [HMDB description or PubChem fallback]

## Reactions (n = N)

Metabolites that directly interact with the input in biochemical reactions.

### Related Metabolite 1

Names: [names]

Description: [description]

**Reaction(s) linking to input:**

- [reaction text]

## Pathways (P pathways, M metabolites)

Pathways providing broader biological context for the input.

### Pathway 1

[pathway description]

**Similar metabolites in this pathway:**

- Names: [names], Description: [description]

## Output Format

[Formatting instructions]

================================================================================

Reaction and pathway sections are included only when relevant context is available. The output format section is always appended at the end of the prompt to encourage adherence during generation.

Non-RAG prompts follow the same structure but include only the Task, Input Metabolite, and Output Format sections, omitting reaction and pathway context.

### B.3.4 Prompt Variants

To increase linguistic diversity during training, we implemented three prompt variants: base, v1, and v2. These variants differed in system prompt phrasing and task instruction wording. Variants were sampled randomly with probabilities 0.50, 0.25, and 0.25 during data generation.

#### B.3.4.1 System Prompt Variants

RAG System Prompts

Base
"You are a biomedical expert in metabolomics and biochemical pathways. Base your reasoning on the provided contexts and integrate general biochemical knowledge."

v1
"As a specialist in metabolomics and biochemical pathways, analyze the input metabolite. Ground your reasoning in the provided contexts while integrating biochemical knowledge."

v2
"You have expertise in biochemical pathways and metabolomics. Reason from the provided evidence and integrate general biochemical knowledge."

Non-RAG System Prompts

Base
"You are a biomedical expert in metabolomics and biochemical pathways. Draw on your biochemical knowledge to describe the input metabolite."

v1
"As a specialist in metabolomics and biochemical pathways, analyze the input metabolite. Use your biochemical knowledge to support your description."

v2
"You have expertise in biochemical pathways and metabolomics. Describe the metabolite based on your biochemical knowledge."

#### B.3.4.2 Task Description Variants

Structure-rich Task Variants

Base

“Write a 300–350 word description covering:

1. The biological role and functional significance of the input metabolite
2. Its connections to other metabolites through the provided reactions and pathways
3. Its overall significance in cellular metabolism”

v1

“Provide a comprehensive description (300–350 words) addressing:

1. The biological role and functional significance of the input metabolite
2. Its connections to other metabolites through the provided reactions and pathways
3. Its overall significance in cellular metabolism”

v2

“Compose a description of approximately 300–350 words that covers:

1. The biological role and functional significance of the input metabolite
2. Its connections to other metabolites through the provided reactions and pathways
3. Its overall significance in cellular metabolism”

Structure-poor Task Variants

Base

“Write a 300–350 word description covering:

1. The biological role and functional significance of the input metabolite
2. Its biochemical properties and known functions
3. Its overall significance in cellular metabolism”

v1

“Provide a comprehensive description (300–350 words) addressing:

1. The biological role and functional significance of the input metabolite
2. Its biochemical properties and known functions

3. Its overall significance in cellular metabolism"

v2

"Compose a description of approximately 300–350 words that covers:

1. The biological role and functional significance of the input metabolite
2. Its biochemical properties and known functions
3. Its overall significance in cellular metabolism"

**B.3.4.3 Output Format Instructions**

RAG Output Format

1. Begin directly with content, with no preamble.
2. Write as flowing paragraphs, with no bullet points or headers.
3. Integrate across contexts rather than summarizing each separately.
4. Explain why relationships exist, not just that they exist.

Non-RAG Output Format

1. Begin directly with content, with no preamble.
2. Write as flowing paragraphs, with no bullet points or headers.
3. Explain the underlying biochemical mechanisms.

**B.3.5 Structure-Based Eligibility and Stratification**

A compound is eligible for retrieval-grounded (structure-rich) description if it has retrievable structure, defined as the union of three conditions: (i) $\geq 2$ pathways and $\geq 1$ reaction (G1), (ii) $\geq 2$ pathways and no reactions (G2), or (iii) $\geq 1$ reaction and no pathways (G3). The remaining compounds form the structure-poor pool, the large majority of the space (224,226 of 237,243; 94.51%). The structure-rich training set was stratified across the three groups (70% G1, 20% G2, 10% G3) so that partially annotated compounds (G2, G3) were represented alongside fully annotated ones (G1), rather than restricting supervision to the richest compounds. The structure-poor set was drawn from the remaining pool without retrieved context.

**Supplementary Table 1.** Description-generation performance measured by ROUGE-L-F1

| Category | Model | Metabolite | Pathway | Enzyme | Struct-rich | Struct-poor |
|---|---|---|---|---|---|---|
| Medical | MedGemma-1.5-4B | 15.41 | 15.27 | 11.19 | 19.07 | 19.58 |
| | FineMedLM-O1 | 16.73 | 16.94 | 12.21 | 18.51 | 18.82 |
| | II-Medical-8B-1706 | 15.81 | 16.76 | 12.04 | 21.53 | 20.96 |
| | Qwen2.5-Aloe-Beta-7B | 16.41 | 17.38 | 12.07 | 19.71 | 18.49 |
| | Meditron3-Qwen2.5-7B | 18.29 | 17.83 | 13.91 | 18.92 | 17.99 |
| Base | Llama-3.2-3B | 16.25 | 17.09 | 12.55 | 19.46 | 19.43 |
| | Gemma-3-4B | 14.93 | 14.80 | 10.88 | 19.09 | 19.74 |
| | Qwen3-4B | 16.56 | 17.12 | 12.78 | 22.11 | 21.72 |
| | Qwen3-8B | 17.62 | 18.27 | 13.41 | 21.95 | 21.57 |
| CPT only | Llama-3.2-3B (CPT) | 23.72 | 21.31 | 13.65 | 16.06 | 17.33 |
| | Gemma-3-4B (CPT) | 25.29 | 18.51 | 11.81 | 17.24 | 17.39 |
| | Qwen3-4B (CPT) | 29.25 | 19.30 | 12.75 | 21.63 | 21.19 |
| | Qwen3-8B (CPT) | 27.53 | 19.39 | 13.98 | 20.49 | 21.16 |
| MetaboLLM | MetaboLLM-Llama-3.2-3B | 54.25 | 31.95 | 20.40 | 23.88 | 27.93 |
| | MetaboLLM-Gemma-3-4B | 39.12 | 23.64 | 13.81 | 23.68 | 27.38 |
| | MetaboLLM-Qwen3-8B | 54.32 | 32.65 | 20.35 | 25.81 | 28.88 |
| | MetaboLLM-Qwen3-4B | **58.65** | **36.71** | **22.28** | **26.42** | **28.96** |

Best per column in bold.

**Supplementary Table 2.** Description-generation performance measured by BLEU-2

| Category | Model | Metabolite | Pathway | Enzyme | Struct-rich | Struct-poor |
|---|---|---|---|---|---|---|
| Medical | MedGemma-1.5-4B | 8.34 | 8.49 | 3.98 | 18.74 | 20.09 |
| | FineMedLM-O1 | 10.79 | 10.32 | 5.31 | 17.47 | 18.23 |
| | II-Medical-8B-1706 | 9.56 | 10.49 | 5.55 | 22.34 | 22.72 |
| | Qwen2.5-Aloe-Beta-7B | 10.61 | 11.28 | 5.66 | 19.25 | 18.85 |
| | Meditron3-Qwen2.5-7B | 11.89 | 11.04 | 6.62 | 16.57 | 17.14 |
| Base | Llama-3.2-3B | 10.03 | 9.98 | 5.56 | 16.88 | 18.13 |
| | Gemma-3-4B | 8.27 | 8.41 | 4.11 | 18.39 | 19.86 |
| | Qwen3-4B | 9.13 | 9.06 | 4.52 | 21.54 | 22.30 |
| | Qwen3-8B | 9.38 | 9.57 | 4.88 | 20.48 | 19.72 |
| CPT only | Llama-3.2-3B (CPT) | 17.58 | 14.66 | 7.37 | 10.71 | 13.98 |
| | Gemma-3-4B (CPT) | 18.65 | 12.25 | 5.40 | 14.31 | 15.95 |
| | Qwen3-4B (CPT) | 22.60 | 11.41 | 5.79 | 20.30 | 20.17 |
| | Qwen3-8B (CPT) | 20.07 | 10.95 | 5.56 | 18.95 | 19.52 |
| MetaboLLM | MetaboLLM-Llama-3.2-3B | 47.37 | 24.74 | 14.11 | 26.63 | 30.58 |
| | MetaboLLM-Gemma-3-4B | 29.62 | 15.48 | 7.03 | 26.70 | 31.43 |
| | MetaboLLM-Qwen3-8B | 47.25 | 25.24 | 13.97 | 28.20 | 31.61 |
| | MetaboLLM-Qwen3-4B | **52.22** | **29.45** | **15.47** | **29.25** | **32.16** |

Best per column in bold.

**Supplementary Table 3.** Pathway enrichment of the top 100 metabolites for stress hyperglycemia prediction

| Pathway | Overlap size / Pathway size | p-value |
|---|---|---|
| Primary bile acid biosynthesis | 11 / 39 | 0.0009 |
| 27-Hydroxylase Deficiency / Cerebrotendinous Xanthomatosis (CTX) / Congenital Bile Acid Synthesis Defect Type II / Congenital Bile Acid Synthesis Defect Type III / Zellweger Syndrome | 10 / 34 | 0.0009 |
| Vitamin A Deficiency | 6 / 14 | 0.0010 |
| Retinol metabolism | 6 / 15 | 0.0012 |
| Metabolism of xenobiotics by cytochrome P450 | 6 / 18 | 0.0038 |
| 11-beta-Hydroxylase Deficiency (CYP11B1) / 17-alpha-Hydroxylase Deficiency (CYP17) / 21-Hydroxylase Deficiency (CYP21) / 3-beta-Hydroxysteroid Dehydrogenase Deficiency / Adrenal Hyperplasia Type 3 or Congenital Adrenal Hyperplasia Due to 21-Hydroxylase Deficiency / Adrenal Hyperplasia Type 5 or Congenital Adrenal Hyperplasia Due to 17 alpha-Hydroxylase Deficiency / Apparent Mineralocorticoid Excess Syndrome / Congenital Lipoid Adrenal Hyperplasia | 6 / 29 | 0.0391 |
| (CLAH) or Lipoid CAH / Corticosterone Methyl Oxidase I Deficiency (CMO I) / Corticosterone Methyl Oxidase II Deficiency (CMO II) | | |
| Steroid hormone biosynthesis | 9 / 53 | 0.0418 |
| Azathioprine Action Pathway / Mercaptopurine Action Pathway / Thioguanine Action Pathway (old) | 8 / 45 | 0.0427 |
| Ubiquinone and other terpenoid-quinone biosynthesis | 4 / 16 | 0.0498 |
| Serotonergic synapse | 6 / 33 | 0.0685 |

**Supplementary Table 4.** Pathway enrichment of the top 100 metabolites for postmenopausal hormone regimen classification

| Pathway | Overlap size / Pathway size | p-value |
|---|---|---|
| ABC transporters | 15 / 44 | 0.0074 |
| Biosynthesis of various other secondary metabolites | 6 / 12 | 0.0122 |
| Amino sugar and nucleotide sugar metabolism | 4 / 6 | 0.0127 |
| 2-Hydroxyglutric Aciduria (D and L Form) / 4-Hydroxybutyric Aciduria/Succinic Semialdehyde Dehydrogenase Deficiency / Homocarnosinosis / Hyperinsulinism-Hyperammonemia Syndrome / Succinic Semialdehyde Dehydrogenase Deficiency | 5 / 9 | 0.0129 |
| Biosynthesis of various plant secondary metabolites | 4 / 7 | 0.0265 |
| Phenylalanine, tyrosine and tryptophan biosynthesis | 4 / 7 | 0.0271 |
| AICA-Ribosiduria / Adenine Phosphoribosyltransferase Deficiency (APRT) / Adenosine Deaminase Deficiency / Adenylosuccinate Lyase Deficiency / Gout or Kelley-Seegmiller Syndrome / Lesch-Nyhan Syndrome (LNS) / Mitochondrial DNA Depletion Syndrome-3 / Molybdenum Cofactor Deficiency / Myoadenylate Deaminase Deficiency / Purine Nucleoside Phosphorylase Deficiency / Xanthine Dehydrogenase Deficiency (Xanthinuria) / Xanthinuria Type I / Xanthinuria Type II | 6 / 14 | 0.0283 |
| D-Amino acid metabolism | 8 / 22 | 0.0297 |
| Diabetic cardiomyopathy | 5 / 11 | 0.0336 |
| Insulin resistance | 5 / 11 | 0.0350 |

**Supplementary Table 5.** Continual pretraining hyperparameters

| Parameter | Value |
|---|---|
| Rank ($r$) | 32 |
| Alpha ($\alpha$) | 64 |
| Base Model | Qwen3-4B-Instruct-2507 |
| Learning Rate | 3e-5 |
| Epochs | 2 |
| Batch Size | 4 |
| Gradient Accumulation | 4 |
| Max Sequence Length | 1536 tokens |
| LR Scheduler | Cosine |
| Optimizer | AdamW |
| Attention | Flash Attention 2 |

**Supplementary Table 6.** Supervised fine-tuning hyperparameters

| Parameter | Value |
|---|---|
| Rank ($r$) | 32 |
| Alpha ($\alpha$) | 64 |
| Dropout | 0.0 |
| Base Model | Qwen3-4B-Instruct-2507 |
| Initialization | Warm-start from CPT checkpoint |
| Learning Rate | 2e-5 |
| Epochs | 3 |
| Batch Size | 2 |
| Gradient Accumulation | 8 |

| Max Sequence Length | 8192 tokens |
|---|---|
| LR Scheduler | Cosine |
| Optimizer | AdamW |
| Attention | Flash Attention 2 |

**Supplementary Table 7.** Retrieval parameters used for retrieval-grounded metabolite description generation

| Parameter | Value |
|---|---|
| Max Reaction Metabolites | 3 |
| Max Pathways | 3 |
| Metabolites per Pathway | 1 |
| Bio Weight ($w_{bio}$) | 0.5 |
| Chem Weight ($w_{chem}$) | 0.5 |
| Duplicate Threshold | 0.98 |

**Supplementary Table 8.** MetaboLLM-GIN hyperparameters and evaluation settings for prediction of stress hyperglycemia after CABG

| Parameter | Value |
|---|---|
| Hidden Dimension | 96 |
| Number of GIN Layers | 2 |
| Dropout Rate | 0.5 |
| Learning Rate | 0.0006 |
| Graph Readout Dimension | 32 |
| Epochs | 50 |
| Batch Size | 8 |

| Weight Decay | 0.01 |
| --- | --- |
| Minimum Learning Rate | 0.00001 |
| Evaluation Scheme | 5-fold cross-validation |

**Supplementary Table 9.** MetaboLLM-GIN hyperparameters and evaluation settings for classification of postmenopausal hormone regimens

| Parameter | Value |
| --- | --- |
| Hidden Dimension | 80 |
| Number of GIN Layers | 3 |
| Dropout Rate | 0.75 |
| Learning Rate | 0.0002 |
| Graph Readout Dimension | 56 |
| Epochs | 50 |
| Batch Size | 16 |
| Weight Decay | 0.008 |
| Minimum Learning Rate | 0.00001 |
| Evaluation Scheme | 80% training and 20% test holdout |